\documentclass[10pt,journal,compsoc]{IEEEtran}
\ifCLASSOPTIONcompsoc
  \usepackage[nocompress]{cite}
\else
  \usepackage{cite}
\fi
\ifCLASSINFOpdf
\else
\fi
\usepackage{times}
\usepackage{epsfig}
\usepackage{graphicx}
\usepackage{caption}
\usepackage{subcaption}
\usepackage{bm}
\usepackage{mathrsfs}
\usepackage{url}
\usepackage{amsmath}
\usepackage{amssymb}
\usepackage{mathtools}
\usepackage{booktabs}

\usepackage{algorithm} 
\usepackage{algorithmic}
\newcommand{\model}[1]{\mathcal{#1}}

\newcommand{\tablequalwidthWACV}{0.09}

\usepackage{array,multirow}
\usepackage{rotating}

\newcommand{\R}{\mathbb{R}}

\newcommand{\VEC}{\operatorname{vec}}

\begin{document}
%
\title{Learning Quadrangulated Patches For \\ 3D Shape Processing}
%
%
%
%

\author{Kripasindhu Sarkar,
        Kiran Varanasi,
        and~ Didier Stricker
\IEEEcompsocitemizethanks{
\IEEEcompsocthanksitem All authors are from DFKI Kaiserslautern
\IEEEcompsocthanksitem K. Sarkar and D. Stricker are from  Technische Universit{\"a}t Kaiserslautern}
}

\IEEEtitleabstractindextext{%
\begin{abstract}
We propose a system for surface completion and inpainting of 3D shapes using generative models, learnt on local patches. 
Our method uses a novel encoding of height map based local patches parameterized using 3D mesh quadrangulation of the low resolution input shape. This provides us sufficient amount of local 3D patches to learn a generative model for the task of repairing moderate sized holes. 
Following the ideas from the recent progress in 2D inpainting, we investigated both linear dictionary based model and convolutional denoising autoencoders based model for the task for inpainting, and show our results to be better than the previous geometry based method of surface inpainting. We validate our method on both synthetic shapes and real world scans.
\end{abstract}

}

\maketitle

\IEEEdisplaynontitleabstractindextext

%
\IEEEpeerreviewmaketitle

\IEEEraisesectionheading{\section{Introduction}\label{sec:introduction}}

%
%
%
%
In recent years, machine learning approaches (CNNs) have achieved the state of the art results for both discriminative \cite{Krizhevsky2012,Girshick2014,Girshick2015,Ren2015,Ren2015} and generative tasks \cite{Mao2016,Cai2017,Pathak2016,Aharon2006, Mairal2008, Mairal2009,Dabov2007,Dabov2009,Lebrun2012,Radford2015,Ledig2016,Pathak2016}.
However, applying the ideas from these powerful learning techniques like Dictionary Learning and Convolutional Neural Networks (CNNs) to 3D shapes is not straightforward, as a common parameterization of the 3D mesh has to be decided before the application of the learning algorithm. A simple way of such parameterization is the voxel representation of the shape. For discriminative tasks, this generic representation of voxels performs very well \cite{Maturana2015, Wu2015, Su2015, Brock2016}. However when this representation is used for global generative tasks, the results are often blotchy, with spurious points floating as noise \cite{Wu2016, Dai2016, Brock2016}. The aforementioned methods reconstruct the global outline of the shape impressively, but smaller sharp features are lost - mostly due to the problem in the voxel based representation and the nature of the problem being solved, than the performance of the CNN.

In this paper we intend to reconstruct fine scale surface details in a 3D shape using ideas taken from the powerful learning methods used in 2D domain. This problem is different from voxel based shape generation where the entire global shape is generated with the loss of fine-scale accuracy. Instead, we intend to restore and inpaint surfaces, when it is already possible to have a global outline of the noisy mesh being reconstructed. 
Instead of the lossy voxel based global representation, we propose local patches computed by the help of mesh quadriangulation. These local patches provide a collection of \textit{fixed-length} and \textit{regular} local units for 3D shapes which can be then used for fine-scale shape processing and analysis.

\begin{figure}[t]
\centering
\begin{subfigure}{1\linewidth}
  \centering
  \includegraphics[width=\linewidth]{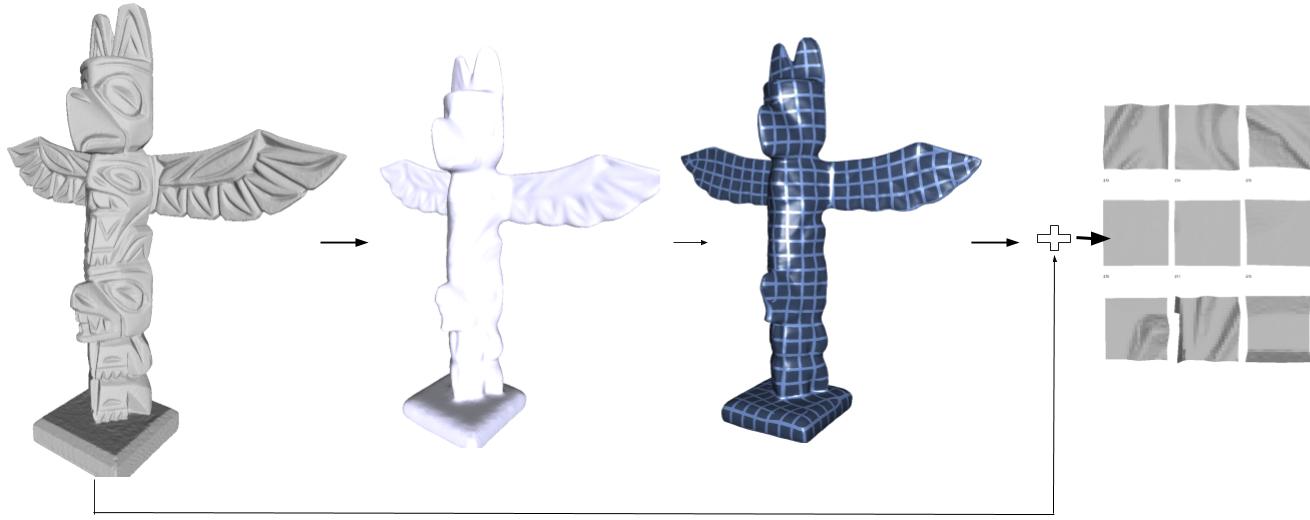}  
\end{subfigure}%
\caption{Our patch computation framework - Local patches are computed on reference frames from quad orientations of the quad mesh obtained from the low resolution version of the input mesh.}
\label{fig:patchframework}
\end{figure}

Our local patch computation procedure makes it possible to have a large number overlapping patches of intermediate length from a single mesh. These patches cover the surface variations of the mesh and are sufficient in amount to use popular machine learning algorithms such as deep CNNs. At the same time due to the stable orientation of our patch computation procedure (by the help of quadriangulations), they are sufficiently large to capture meaningful surface details. This makes these patches suitable for the application of repairing a damaged part in the same mesh, while learning from its undamaged parts or some other clean meshes. Because of the locality and the density of the computed patches, we do not need a large database of shapes to correct a damaged part or fill a moderate sized hole in a mesh. We explore ideas from 2D images and use methods such as dictionary learning and deep generative CNNs for surface analysis.

We compute local patches of moderate length by applying automatic mesh quadrangulation algorithm~\cite{Ebke2013} to the low-resolution representation of an input 3D mesh and taking the stable quad orientations for patch computation. The low-resolution mesh is obtained by applying mesh smoothing to the input 3D scan, which captures the broad outline of the shape over which local patches can be placed. We then set the average quad size and thereby choose the required scale for computing local patches. The mesh quadrangulation is a quasi-global method, which determines the local orientations of the quads based on the distribution of corners and edges on the overall shape. At the same time, the scanlines of the quads retain some robustness towards surface noise and partial scans - these orientations can be fine-tuned further by the user if needed. The patch computation approach is summarized in Figure \ref{fig:patchframework}.

Prior work in using local surface patches for 3D shape compression \cite{Digne2014} assumed the patch size to be sufficiently small such that a local patch on the input 3D scan could be mapped to a unit disc. Such small patch sizes would restrict learning based methods from learning any shape detail at larger scale, and for applications like surface inpainting. 

The contributions of our paper are as follows. 
\begin{enumerate}
\item We propose a novel shape encoding by local patches oriented by mesh quadrangulation. Unlike previous works, we do not require
the patches to be exceedingly small \cite{Digne2014}.

\item Using our quadriangulated patches, we propose a method for learning a 3D patch dictionary. Using the self-similarity among the 3D patches we solve the problem of surface analysis such as inpainting and compression.

\item We extend the insights for designing CNN architectures for 2D image inpainting to surface inpainting of 3D shapes using our 3D patches. We provide analysis for their applicability to shape denoising and inpainting.

\item We validate the applicability of our models (patch-dictionary and CNN) learned from multiple 3D scans thrown into a common data set, towards repairing an individual 3D scan. 
\end{enumerate}

The related work is discussed in the following section. We first explain our encoding of quadriangulated patches in Section \ref{sec:3Dpatches}. We then present both linear and CNN based generative models in Section \ref{sec:generativemodels}. We follow it with the experiments section where both the generative models are evaluated.

\section{Related Work}
\label{sec:related_work}
\subsection{3D global shape parameterization} Aligning a data set of 3D meshes to a common global surface parameterization is very challenging and requires the shapes to be of the same topology. For example, {\em geometry images}\cite{Sinha2016} can parameterize genus-0 shapes on a unit sphere, and even higher topology shapes with some distortion. Alternatively, the shapes can be aligned on the spectral distribution spanned by the Laplace-Beltrami Eigenfunctions\cite{Masci2015a,Boscaini2016}. However, even small changes to the 3D mesh structure and topology can create large variations in the global spectral parameterization - something which cannot be avoided when dealing with real world 3D scans. Another problem is with learning partial scans and shape variations, where the shape detail is preserved only locally at certain places. Sumner and Popovic \cite{Sumner2004} proposed the {\em deformation gradient} encoding of a deforming surface through the individual geometric transformations of the mesh facets. This encoding can be used for statistical modeling of pre-registered 3D scans~\cite{Neumann2013}, and describes a Riemannian manifold structure with a Lie algebra~\cite{Freifeld2012}. All these methods assume that the shapes are pre-registered globally to a common mesh template, which is a significant challenge for shapes with arbitrary topologies. Another alternative is to embed a shape of arbitrary topology in a set of 3D cubes in the extrinsic space, known as {\em PolyCube-Maps}~\cite{Tarini2004}. Unfortunately, this encoding is not robust to intrinsic deformations of the shape, such as bending and articulated deformations that can typically occur with real world shapes. So we choose an intrinsic quadrangular parameterization on the shape itself~\cite{Ebke2013}(see also Jakob et al.~\cite{Jakob2015}).

\subsection{Statistical learning of 3D shapes} For reconstructing specific classes of shapes, such as human bodies or faces, fine-scale surface detail can be learned~{\em e.g,}\cite{Garrido2016,Bermano2014,Bogo2015}, from high resolution scans registered to a common mesh template model. This presumes a common shape topology or registration to a common template model, which is not possible for arbitrary shapes as presented in our work. 
For shapes of arbitrary topology, existing learning architectures for deep neural networks on 2D images can be harnessed by using the projection of the model into different perspectives~\cite{Su2015, Sarkar2017}, or by using its depth images~\cite{Wei2016}. 3D shapes are also converted into common global descriptors by voxel sampling. The availability of large database of 3D shapes like ShapeNet \cite{Chang2015} has made possible to learn deep CNNs on such voxalized space for the purpose of both discrimination \cite{Maturana2015, Wu2015, Su2015, Brock2016} and shape generation \cite{Wu2016, Dai2016, Brock2016}. Unfortunately, these methods cannot preserve fine-scale surface detail, though they are good for identifying global shape outline. More recently, there has been serious effort to have alternative ways of applying CNNs in 3D data such as OctNet \cite{Riegler2017} and PointNet \cite{Qi2016}. OctNet system uses a compact version of voxel based representation where only occupied grids are stored in an octree instead of the entire voxel grid, and has similar computational power as the voxel based CNNs. 
PointNet on the other hand takes unstructured 3D points as input and gets a global feature by using max pool as a symmetrical function on the output of MLP (multi-layer perceptron) on individual points. 
Both these networks have not been explored yet fully for their generation properties (Eg. OctNetFusion \cite{Riegler2017a}). They are still in their core, systems for global representation and are not targeted specifically for surfaces. In contrast, we encode 3D shape by fixed-length and regular local patches and learn generative models (patch dictionary and generative CNNs) for reproducing fine scaled surface details.

\subsection{CNN based generative models in images} One of the earliest work on unsupervised feature learning are autoencoders \cite{Hinton2006} which can be also seen as a generative network. A slight variation, denoising autoencoders \cite{Vincent2008,Xie2012}, reconstruct the image from local corruptions, and are used as a tool for both unsupervised feature learning and the application of noise removal. Our generative CNN model is, in principle, a variant of denoising autoencoder, where we use convolutional layers following the modern advances in the field of CNNs. \cite{Mao2016,Cai2017,Pathak2016} uses similar network with convolutional layers for image inpainting. Generating natural images from using a neural network has also been studied extensively - mostly after the introduction of Generative Adversarial Network (GAN) by Goodfellow \cite{Goodfellow2014} and its successful implementation using convolutional layers in DCGAN (Deep Convolutional GANs) \cite{Radford2015}. As discussed in Section \ref{sec:networkdesign}, our networks for patch inpainting are inspired from all the aforementioned ideas and are used to inpaint height map based 3D patches instead of images.

\subsection{Dense patch based generative models in images} 2D patch based methods have been very popular in the topic of image denoising. These non local algorithms can be categorised into dictionary based \cite{Aharon2006, Mairal2008, Mairal2009} and BM3D (Block-matching and 3D filtering) based \cite{Dabov2007, Dabov2009, Lebrun2012} methods. 
Because of the presence of a block matching step in BM3D (patches are matched and kept in a block if they are similar), it is not simple to extend it for the task of inpainting, though the algorithm can be applied indirectly in a different domain \cite{Li2014}. In contrast, dictionary based methods can be extended for the problem of inpatinting by introducing missing data masks in the matrix factorization step - making them the most popular methods for the comparison of inpainting tasks. 
In 3D meshes, due to the lack of common patch parameterization, this task becomes difficult. 
In this work, we use our novel encoding to compute moderate length dense 3D patches, and process them with the generative models of patch dictionary and non linear deep CNNs.

\subsection{3D patch dictionaries} A lossy encoding of local shape detail can be obtained by 3D feature descriptors~\cite{Kim2013}. However, they typically do not provide a complete local surface parameterization. Recently, Digne et al.~\cite{Digne2014} used a 3D patch dictionary for point cloud compression. Local surface patches are encoded as 2D height maps from a circular disc and learned a sparse linear dictionary of patch variations~\cite{Aharon2006}. They assume that the local patches are sufficiently small (wherethe shape is parameterizable to a unit disc). In contrast to
this work, (i) we use mesh quadringulation for getting the patch location and orientation (in comparison to uniform sampling and PCA in \cite{Digne2014}) enabling us to get large patches at good locations, (ii) 
we address the problem of inpainting by generative models (masked version in matrix factorization and a blind method for CNN models) instead of compression, (iii) as a result of aforementioned differences, our patch size is much larger in order to have a meaningful patch description in the presence of missing regions.

\subsection{General 3D surface inpainting} Earlier methods for 3D surface inpainting regularized from the geometric neighborhood ~\cite{Liepa2003,Bendels2006}. More recently, Sahay et al.~\cite{Sahay2015} inpaint the holes in a shape by pre-registering it to a {\em self-similar} proxy model in a dataset, that broadly resembles the shape. The holes are inpainted using a patch-dictionary. In this paper, we use a similar approach, but avoid the assumption of finding and pre-registering to a proxy model. The term {\em self-similarity} in our paper refers to finding similar patches in other areas of the shape. Our method automatically detects the suitable patches, either from within the shape, or from a diverse data set of 3D models. Zhong et al.~\cite{Zhong2016} propose an alternative learning approach by applying sparsity on the Laplacian Eigenbasis of the shape. We show that our method (both patch dictionary and generative CNN models) is better than this approach on publicly available meshes.

\begin{figure}[t]
\centering
\begin{subfigure}{0.6\linewidth}
  \centering
  \includegraphics[width=0.8\linewidth]{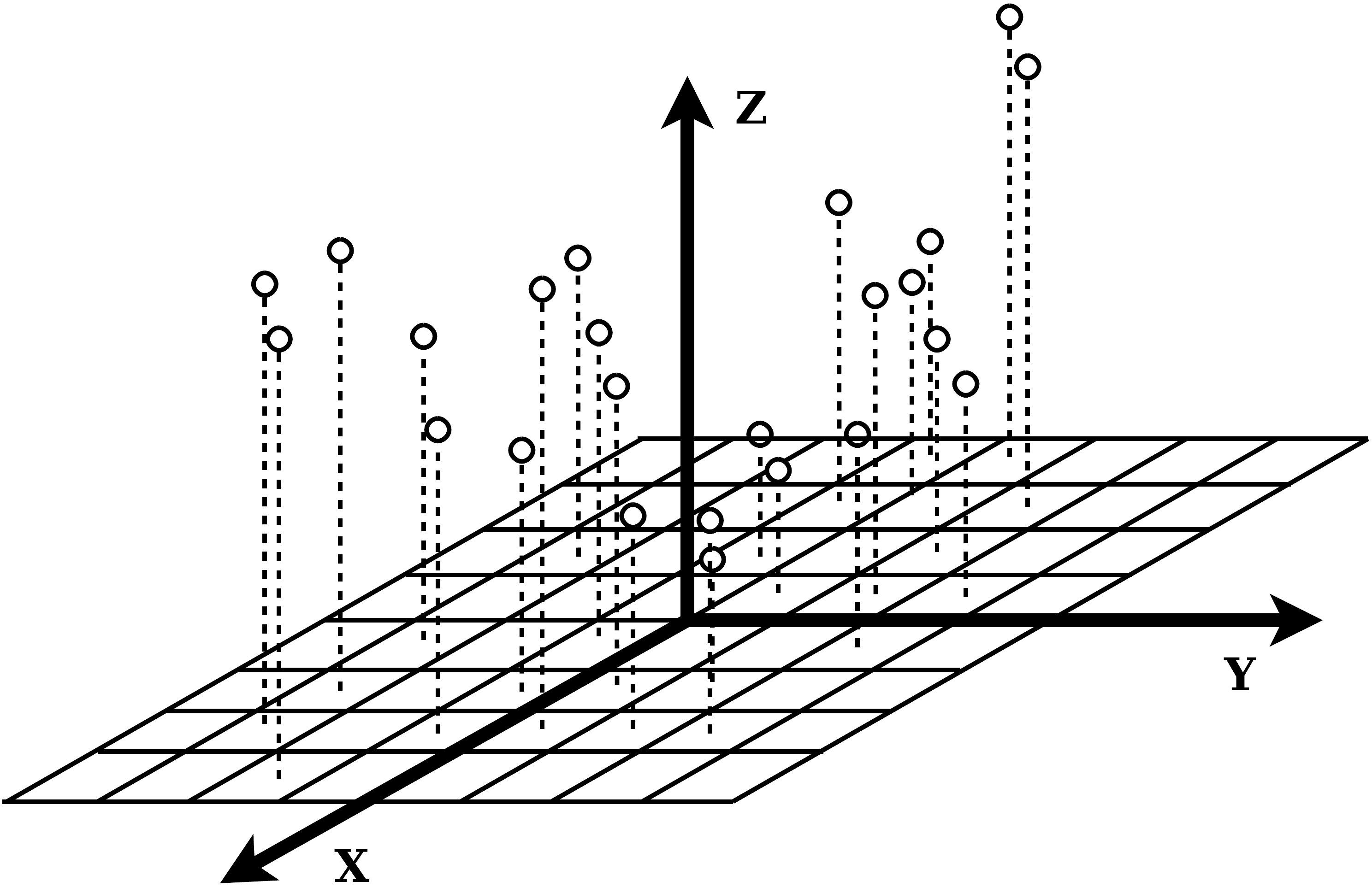}  
\end{subfigure}%
\begin{subfigure}{0.35\linewidth}
  \centering
  \includegraphics[width=0.8\linewidth]{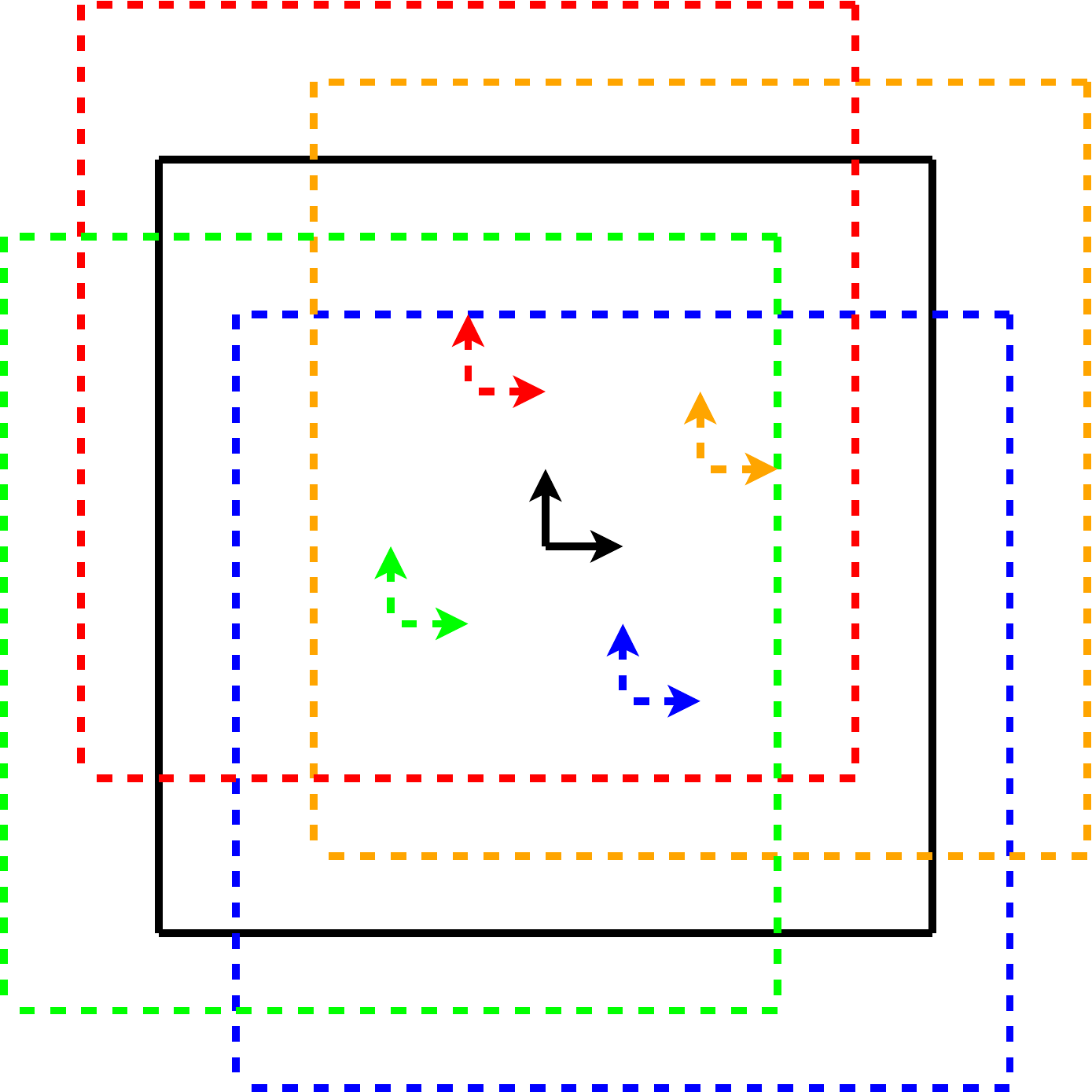}  
\end{subfigure}
\caption{(Left) Patch representation - Points are sampled as a height map over the planer grid of a reference frame at the seed point. (Right) Patches computed at multiple offset from the quad centres to simulate dense sampling of patches while keeping the stable quad orientation. The black connected square represents the quad in a quad mesh and the dotted squares represents the patches that are computed at different offset.}
\label{fig:heightmap}
\end{figure}

\section{3D Patch Encoding}
\label{sec:3Dpatches}

Given a mesh $\model{M} = \{F, V\}$ depicting a 3D shape and the input parameters - patch radius $r$ and grid resolution $N$, our aim is to decompose it into a set of fixed length local patches $\{P_s\}$, along with the settings $\model{S} = \{(s, T_{s})\}, Conn$ having information on the location (by $s$), orientation (by the transformation $T_{s}$) of each patch and vertex connectivity (by $Conn$) for reconstructing back the original shape.

To compute uniform length patches, a point cloud $C$ is computed by dense uniform sampling of points in $\model{M}$. Given a seed point $s$ on the model surface $C$, a reference frame $\model{F}_s$ corresponding to a transformation matrix $T_{s}$ at $s$, and an input patch-radius $r$,  we consider all the points in the $r$-neighbourhood, $\model{P}_s$.  
Each point in $\model{P}_s$ is represented w.r.t. $\model{F}_s$ as $P_{\model{F}_s}$. That is, if the rotation between global coordinates and $\model{F}_s$ is given by the rotation matrix $R_s$, a point $\bm{p}$ represented in the local coordinate system of $\model{F}_s$ is given by $\bm{p}_{s}= T_s \bm{p}$, where $T_s = \begin{pmatrix}R & {-R_s}s\\ 0 & 1\end{pmatrix}$ is the transformation matrix between the two coordinates.

\subsection{Local parameterisation and patch representation}
An $N\times N$ square grid of length $\sqrt{2}r$ and is placed on the X-Y plane of $\model{F}_s$, and points in $P_{\model{F}_s}$ are sampled over the grid wrt their X-Y coordinates. Each sampled point is then represented by its `height' from the square grid, which is its Z coordinate to finally get a height-map representation of dimension of $(N \times N)$ (Figure \ref{fig:heightmap}). Thus, each patch around a point $s$ is defined by a \textit{fixed size} vector $\VEC(P_s)$ of size $N^2$ and a transformation $T_s$. 

\subsection{Mesh reconstruction}
\label{sec:connectedmeshrec}
 To reconstruct a connected mesh from patch set we need to store connectivity information $Conn$. This can be achieved by keeping track of the exact patch-bin $(P_s, i)$ a vertex  $v_j \in V$ in the input mesh corresponds (would get sampled during the patch computation) by the mapping $\{(j, \{(P_s, i)\})\}$.

Therefore, given patch set  $\{P_s\}$ along with the settings $\model{S} = \{(s, T_{s})\}, Conn$ with $Conn = \{(j, \{(P_s, i)\})\}, F$ it is possible to reconstruct back the original shape with the accuracy upto the sampling length. For each patch $P_{s}$, for each bin $i$, the height map representation $P_s[i]$, is first converted to the XYZ coordinates in its reference frame, $\bm{p}_s$, and then to the global coordinates $\bm{p}'$, by $\bm{p}'= T_s^{-1} \bm{p}_s$. Then the estimate of each vertex index $j$, $v_j \in V$ is given by the set of vertices $\{v_e\}$. The final value of vertex $v_m'$ is taken as the mean of $\{v_e\}$. The reconstructed mesh is then given by $\{\{v_j'\}, F\}$. If the estimate of a vertex $v_j$ is empty, we take the average of the vertices in its 1-neighbour ring.

 \begin{algorithm}[t]
\renewcommand{\algorithmicrequire}{\textbf{Input:}}
\renewcommand{\algorithmicensure}{\textbf{Output:}}
\floatname{algorithm}{Steps}
        \caption{3D Patch computation based on quad mesh}
        \begin{algorithmic}[1]
            \REQUIRE Mesh - $M$, Patch radius - $r$, resolution - $N$
            \STATE Compute quad mesh of the smoothened $M$ using \cite{Jakob2015}.
			\STATE Densely sample points in $M$ to get the cloud $C$.
\STATE At each quad center, compute r-neighborhood in $C$ and orient using the quad orientation to get local patches.
\STATE Sample the local patches in a ($N \times N$) square grid in a height map based representation.
\STATE Store the vertex connections (details in the text).
            \ENSURE Patch set $\{P_s\}$ of ($N \times N$) dimension,  orientations, vertex connections.
        \end{algorithmic}
        \label{algorithm}
        
\end{algorithm}

\subsection{Reference frames from quad mesh}
\label{sec:rfcomputation} \label{sec:globalproperties}
The height map based representation accurately encodes a surface only when the patch radius is below the distance between surface points and the shape medial axis. In other words, the $r$-neighbourhood, $\model{P}_s$ should delimit a topological disk on the underlying surface to enable parameterization over the grid defined by the reference frame. In real world shapes, either this assumption breaks, or the patch radius becomes too low to have a meaningful sampling of shape description. A good choice of seed points enables the computation of the patches in well behaved areas, such that, even with moderately sized patches in arbitrary real world shapes, the $r$-neighbourhood, $\model{P}_s$ of a given point $s$ delimits a topological disk on the grid of parameterisation. It should also provide an orientation consistent with global shape. 


Given a mesh $\model{M}$, we obtain low-resolution representation by Laplacian smoothing \cite{Sorkine2004}. The low resolution mesh captures the broad outline of the shape over which local patches can be placed. In our experiments, for all the meshes, we performed $30$  Laplacian smoothing iterations (normal smoothing + vertex fitting).

Given the smooth coarse mesh, the quad mesh $\model{M}^Q$ is extracted following Jakob et al.\cite{Jakob2015}. At this step, the quad length is specified in proportion to the final patch length and hence the scale of the patch computation. For each quad $q$ in the quad mesh, its center and $4*k$ offsets are considered as seed points, where $k$ is the overlap level (Figure \ref{fig:heightmap} (Right)). These offsets capture more patch variations for the learning algorithm. For all these seed points, the reference frames are taken from the orientation of the quad $q$ denoted by its transformation $T_{s}$. In this reference frame, $Z$ axis, on which the height map is computed, is taken to be in the direction normal to the quad. The other two orthogonal axes - $X$ and $Y$, are computed from the two consistent sides of the quads. To keep the orientation of $X$ and $Y$ axes consistent, we do a breath first traversal starting from a specific quad location in the quad mesh and reorient all the axes to the initial axes.

\begin{figure}[t]
\centering
\begin{subfigure}{\linewidth}
  \centering
  \includegraphics[width=\linewidth]{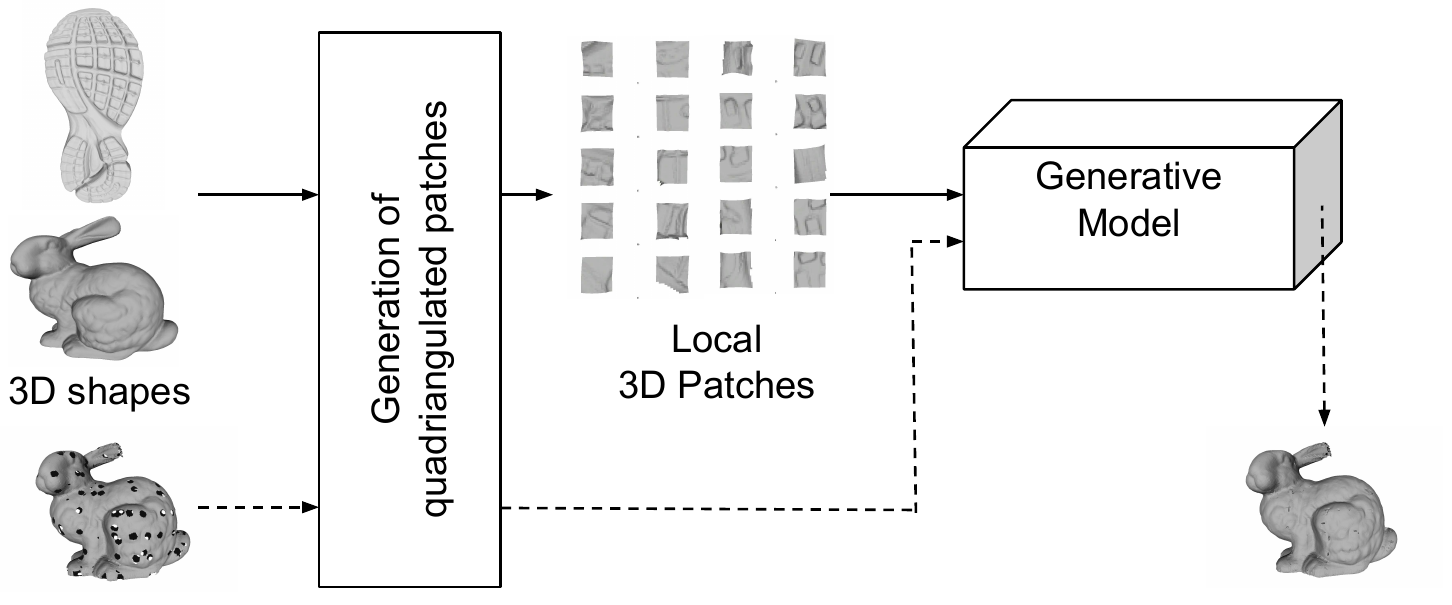}  
\end{subfigure}%
\caption{Summary of the inpainting framework. Generative models are trained on 3D the patches computed from 3D shapes for the purpose of inpainting. During testing (dashed line) the generative model is used to reconstruct noisy patches computed in the noisy mesh.}
\label{fig:patchaa_summary}
\end{figure}

\section{Learning on 3D patches}
\label{sec:generativemodels}
Given a set of 3D meshes, we first decompose them into local rectangular patches. Using this large database of 3D patches, we learn a generative model to reconstruct denoised version of input 3D patches. We use both Matrix Factorization and CNN based generative models for inpainting whose details are explained in this section. The overall approach for training is presented in Figure \ref{fig:patchaa_summary}.

Let $\bm{x_i} := \VEC(P_i) \in \R^{N^2}$ be the vectorization of the patch $P_i$ in the patch set $\{P_i\}$. And let $X$ be the set of the domain of the vectorization of the patches generated from a mesh (or a pool of meshes). Given such patch set, we learn a generative model $\model{M}: X \mapsto X$, such that $\model{M}(\bm{x}) = \bm{x'}$ produces a cleaned version of the noisy input $\bm{x}$. Following sections describe two such popular methods of generative models used in the context of patch inpainting, namely Dictionary Learning and Denoising Autoencoders. These methods, inspired from their popularity in the 2D domain as generative models, are designed to meet the needs of the patch encoding. They are described in detail in the following paragraphs.

\subsection{Dictionary Learning and Sparse Models}
Given a matrix ${D}$ in $\mathbb{R}^{m \times p}$ with $p$ column vectors, sparse models in signal processing aims at representing a signal $\bm{x}$ in $\mathbb{R}^{m}$ as a sparse linear combination of the column vectors of $D$. The matrix $D$ is called \textit{dictionary} and its columns \textit{atoms}. In terms of optimization, approximating $\bm{x}$ by a sparse linear combination of atoms can be formulated as finding a sparse vector $\bm{y}$ in $\mathbb{R}^p$, with $k$ non-zero coefficients, that minimizes
\vspace{-0.2cm}
\begin{equation} \label{eq:sparsity}
  \quad  \min \limits _{\bm{y}} \frac{1}{2}\|\bm{x} - D\bm{y}\|^2_2 \qquad  \text{s.t. } \|\bm{y}\|_0 \le k
    \vspace{-0.2cm}
\end{equation}

The dictionary $D$ can be learned or evaluated from the signal dataset itself which gives better performance over the off-the-shelf dictionaries in natural images. In this work we learn the dictionary from the 3D patches for the purpose of mesh processing. Given a dataset of $n$ training signals $\bm{X} = [\bm{x}_1, ..., \bm{x}_n]$, dictionary learning can be formulated as the following minimization problem
\vspace{-0.2cm}
\begin{equation} \label{eq:dlearning}
  \quad  \min \limits _{D, \bm{Y}} \sum_{i=1}^n \frac{1}{2}\|\bm{x}_i - D\bm{y}_i\|^2_2 + \lambda \psi(\bm{y}_i),
    \vspace{-0.2cm}
\end{equation}

where $\bm{Y} = [\bm{y}_1, ..., \bm{y}_n] \in \mathbb{R}^{p \times n}$ is the set of sparse decomposition coefficients of the input signals $\bm{X}$, $\psi$ is sparsity inducing regularization function, which is often the $l_1$ or $l_0$ norm.

Both optimization problems described by equations \ref{eq:sparsity} and \ref{eq:dlearning} are solved by approximate or greedy algorithms; for example, Orthogonal Matching Pursuit (OMP) \cite{Pati1993}, Least Angle Regression (LARS) \cite{Efron2004} for sparse encoding (optimization of Equation \ref{eq:sparsity}) and KSVD \cite{Aharon2006} for dictionary learning (optimization of Equation \ref{eq:dlearning})

\textbf{Missing Data:} Missing information in the original signal can be well handled by the sparse encoding. To deal with unobserved information, the sparse encoding formulation of Equation \ref{eq:sparsity} can be modified by introducing a binary mask $M$ for each signal $\bm{x}$. Formally, $M$ is defined as a diagonal matrix in $\mathbb{R}^{m \times m}$ whose value on the $j$-th entry of the diagonal is 1 if the pixel $\bm{x}$ is observed and 0 otherwise. Then the sparse encoding formulation becomes 

\begin{equation}\label{eq:maskedsparsity}
  \quad  \min \limits _{\bm{y}} \frac{1}{2}\|M(\bm{x} - D\bm{y})\|^2_2 \qquad  \text{s.t. } \|\bm{y}\|_0 \le k
  \vspace{-0.2cm}
\end{equation}

Here $M\bm{x}$ represents the observed data of the signal $\bm{x}$ and $\bm{x'} = D\bm{y}$ is the estimate of the full signal. The binary mask does not drastically change the optimization procedure and one can still use the classical optimization techniques for sparse encoding.

\subsubsection{3D Patch Dictionary}
\label{sec:shapeencoding}
We learn patch dictionary $D$ with the generated patch set $\{P_s\}$ as training signals ($m = N^2$). This patch set may come from a single mesh (providing \textit{local dictionary}), or be accumulated globally using patches coming from different shapes (providing a \textit{global dictionary} of the dataset). Also in the case of the application of hole-filling, a dictionary can be learnt on the patches from clean part of the mesh, which we call \textit{self-similar} dictionary which are powerful in meshes with repetitive structures. For example a tiled floor, or the side of a shoe has many repetitive elements that can be learned automatically. We computed patches at the resolution of (24 $\times$ 24) following the mesh resolution. More details on the 3D dataset, patch size, resolutions for different types of meshes are provided in the Evaluation section. Please note that, we also computed patches at the resolution (100 $\times$ 100) for longer CNN base generative models they are more complex than the linear dictionary based models. Please find the details in the next section.

\textbf{Reconstruction}
Using a given patch dictionary $D$, we can reconstruct the original shape whose accuracy  depends on the number of atoms chosen for the dictionary. For each 3D patch $\bm{x_i} = \VEC(\bm{P}_i)$ from the generated patches and the learnt dictionary $D$ of a shape, its sparse representation, $\bm{y}$ is found following the optimization in Equation \ref{eq:sparsity} using the algorithm of Orthogonal Matching Pursuit (OMP). It's approximate representation, the locally reconstructed patch $\bm{x}_i'$ is found as $\bm{x}_i' \approx D\bm{y}$. The final reconstruction is performed using the altered patch set $\{P_i'\}$ and $\model{S}$ following the procedure in Section \ref{sec:connectedmeshrec}. 

\textbf{Reconstruction with missing data}
\label{sec:missingdatarec}
In case of 3D mesh with missing data, for each 3D patch $\bm{x_i}$ computed from the noisy data having missing values, we find the sparse encoding $\bm{y_i}$ following Equation \ref{eq:maskedsparsity}. The estimate of the full reconstructed patch is then $\bm{x}' = D\bm{y}$. 

Results of inpainting using Dictionary Learning is provided in the Evaluation section (Section \ref{sec:results}). We now present the second generative model in the next section.

\begin{figure*}
\small
\centering
\begin{subfigure}{0.45\linewidth}
  \centering
  \includegraphics[width=\linewidth]{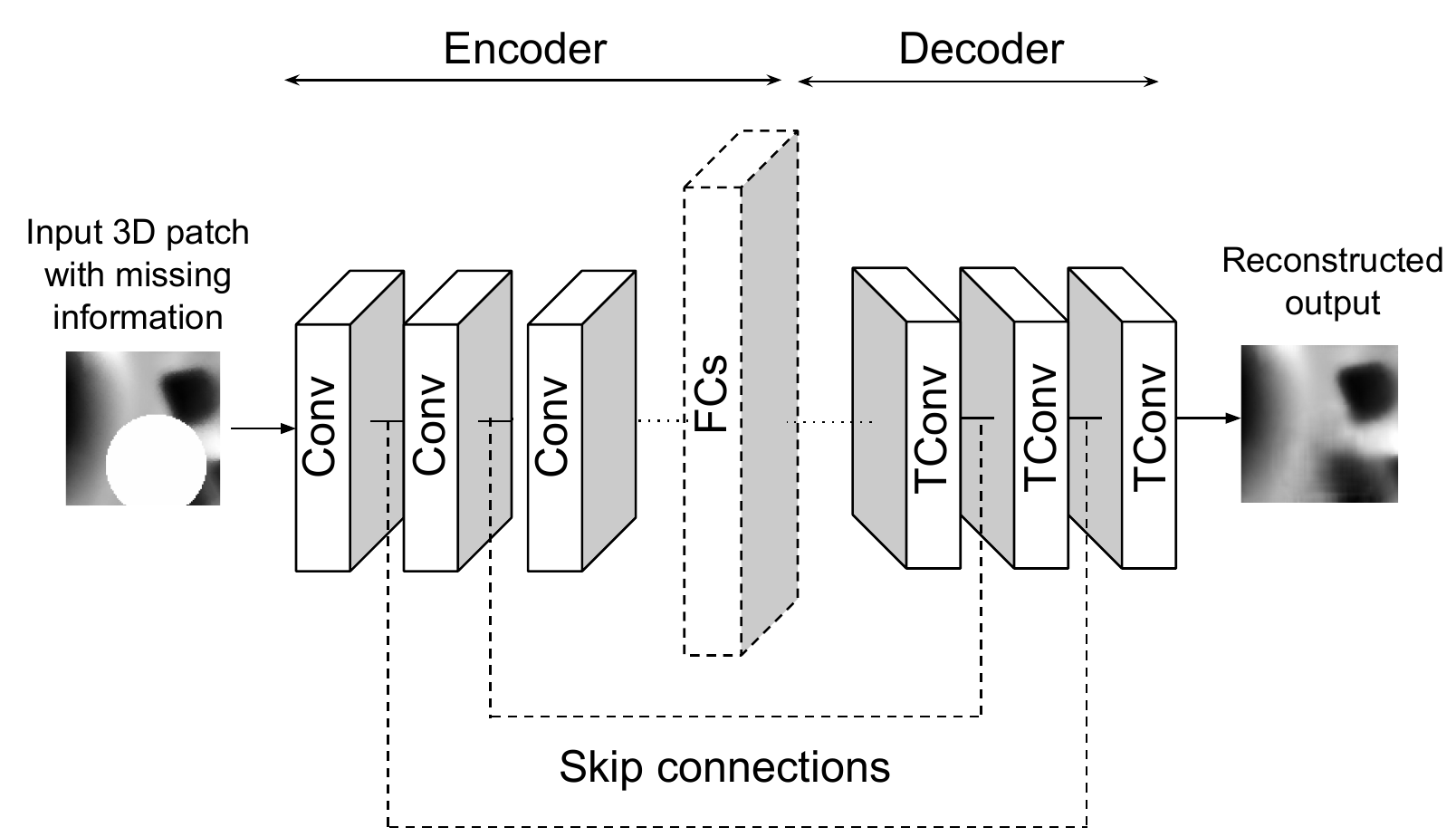}
  \end{subfigure}
\resizebox{0.8\textwidth}{!}{
\begin{tabular}{|l|l|l|l|l|l|l|}
\hline
 & \textbf{small\_4x} & \textbf{multi\_6x} &  \textbf{6x\_128} & \textbf{6x\_128\_FC} & \textbf{long\_12x} & \textbf{long\_12x\_SC} \\ \hline
Input & (24x24) & (100x100x1) & (100x100x1) & (100x100x1) & (100x100x1) & (100x100x1) \\ \hline
 & 3x3, 32 & 3x3, 32 &   &  &  &  \\ 
 & 3x3, 32, (2, 2) & 3x3, 32, (2, 2) &  5x5, 32, (2, 2) & 5x5, 32, (2, 2) & 5x5, 64 & 5x5, 64 \\ \cline{2-7}
 \multirow{3}{*}{\begin{turn}{90} Convolution blocks\end{turn}}  & 3x3, 32 & 3x3, 32 &  &  &  & 5x5, 64, (2, 2) \\ 
 & 3x3, 32, (2, 2) & 3x3, 32, (2, 2) & 5x5, 64, (2, 2) & 5x5, 64, (2, 2) & 5x5, 64, (2, 2) & Out (1) \\ \cline{3-7}
 &  & 3x3, 32 &  &   &  &  \\ 
 &  & 3x3, 32, (2, 2) & 5x5, 128, (2, 2) & 5x5, 128, (2, 2) & 5x5, 64 & 5x5, 64 \\ \cline{5-6} \cline{6-7}

 &  &  &  &  &  & 5x5, 64, (2, 2) \\ 
 &  &  &  &  & 5x5, 64, (2, 2) & Out (2) \\ \cline{6-7}
 &  &  &  &  &  &  \\ 
 &  &  &  &  & 5x5, 64 & 5x5, 64 \\ \cline{6-7}
 &  &  &  &  &  &  \\ 
 &  &  &  & FC 4096  & 5x5, 64, (2, 2) & 5x5, 64, (2, 2) \\ \hline \hline

 &  &  &  &  &  &  \\ 
 &  &  &  &  & 5x5, 64 & 5x5, 64 \\ \cline{6-7}
  \multirow{3}{*}{\begin{turn}{90}Transposed conv blocks\end{turn}} &  &  &  &  &  & 5x5, 64, (2, 2) \\ 
 &  &  &  &  & 5x5, 64, (2, 2) & Relu + (2) \\ \cline{6-7}
 &  & 3x3, 32 &  &   &  &  \\ 
 &  & 3x3, 32, (2, 2) &  5x5, 128, (2, 2) & 5x5, 128, (2, 2)  & 5x5, 64 & 5x5, 64 \\ \cline{3-7}
 & 3x3, 32 & 3x3, 32 &  &  &  & 5x5, 64, (2, 2) \\ 
 & 3x3, 32, (2, 2) & 3x3, 32, (2, 2) &  5x5, 64, (2, 2) & 5x5, 64, (2, 2) & 5x5, 64, (2, 2) & Relu + (1) \\ \cline{2-7}
 & 3x3, 32 & 3x3, 32 &  &  &  &  \\ 
 & 3x3, 32, (2, 2) & 3x3, 32, (2, 2) & 5x5, 32, (2, 2) & 5x5, 32, (2, 2)  & 5x5, 64 & 5x5, 64 \\ \cline{2-7}
 
 & 3x3, 1 & 3x3, 1 & 5x5, 1 & 5x5, 1  & 5x5, 1, (2, 2) & 5x5, 1, (2, 2) \\ \hline
\end{tabular}

}
\caption{(Left) - Summary of our network architecture showing the building blocks. Dashed lines and blocks are optional parts depending on the network as described in the table on the right. Conv, FCs and TConv denote Convolution, Fully Connected and Transposed Convolution layers respectively. (Right) - The detailed description of the different networks used. Each column represents a network where the input is processed from top to bottom. The block represents the kernel size, number of filters or output channels and optional strides when it differs from (1, 1). The network complexity in terms of computation and parameters increases from left to right except for \textit{6x\_128\_FC}, which has the maximum number of parameters because of the presence of the FC layer. Other details are provided in Section \ref{sec:networkdesign}.}
\label{table:networks}
\end{figure*}
    
\subsection{Denoising Autoencoders for 3D patches}
\label{sec:cnnintro}
In this section we present the generative model $\model{M}: X \mapsto X$ as Convolutional Denoising Autoencoder. Autoencoders are generative networks which try to reconstruct the input. A Denoising Autoencoder reconstructs the de-noised version of the noisy input, and is one of the most well known method for image restoration and unsupervised feature learning \cite{Xie2012}. We use denoising autoencoder architecture with convolutional layers following the success of general deep convolutional neural networks (CNN) in images classification and generation. Instead of images, we use the 3D patches generated from different shapes as input, and show that this height map based representation can be successfully used in CNN for geometry restoration and surface inpainting. 

Following typical denoising autoencoders, our network has two parts - an encoder and a decoder. An encoder takes a 3D patch with missing data as input and and produces a latent feature representation of that image. The decoder takes this feature representation and reconstructs the original patch with missing content. The encoder contains a sequence of convolutional layers which reduces the spatial dimension of the output as we go forward the network. Therefore, this part can be also called \textit{downsampling} part. This follows by an optional fully connected layer completing the encoding part of the network. The decoding part consists fractionally strided convolution (or transposed convolution) layers which increase the spatial dimension back to the original patch size and hence can also be called as \textit{upsampling}. The general design is shown in Figure \ref{table:networks} (Left).

\subsubsection{Network design choices}
\label{sec:networkdesign}
Our denoising autoencoder should be designed to meet the need of the patch encoding. The common design choices are presented in Figure \ref{table:networks} and are discussed in the following paragraphs in details.

\textbf{Pooling vs strides} 
Following the approach of powerful generative models like Deep Convolutional Generative Adversarial Network (DCGAN) \cite{Radford2015}, we use strided convolutions for downsampling and strided transposed convolutions for upsampling and do not use any pooling layers. For small networks its effect is insignificant, but for large network the strided version performs better.

\textbf{Patch dimension} We computed patches at the resolution of 16 $\times$ 16, 24 $\times$ 24 and 100 $\times$ 100 with the same patch radius (providing patches at the same scale) in our 3D models. Patches with high resolution capture more details than the low resolution counterpart. But, reconstructing higher dimension images is also difficult by a neural network. This causes a trade-off which needs to be considered. Also higher resolution requires a bigger network to capture intricate details which is discussed in the following paragraphs. For lower dimensions (24 $\times$ 24 input), we used two down-sampling blocks followed by two up-sampling blocks. We call this network \textbf{small\_4x} as described in Figure \ref{table:networks}. 
Other than this, all the considered network take an input of 100 $\times$ 100 dimensions. The simplest ones corresponding to 3 encoder and decoder blocks are  \textbf{multi\_6x} and \textbf{6x\_128}.

\textbf{Kernal size} Convolutional kernel of large size tends to perform better than lower ones for image inpainting. \cite{Mao2016} found a filter size of (5 $\times$ 5) to (7 $\times$ 7) to be the optimal and going higher degrades the quality. Following this intuition and the general network of DCGAN \cite{Radford2015}, we use filter size of (5 $\times$ 5) in all the experiments.

\textbf{FC latent layer} A fully connected (FC) layer can be present in the end of encoder part. If not, the propagation of information from one corner of the feature map to other is not possible. However, adding FC layer where the latent feature dimension from the convolutional layer is already high, will cause explosion in the number of parameters. It is to be noted that for inpainting, we want to retain as much of information as possible, unlike simple Autoencoders where the latent layer is often small for compact feature representation and dimension reduction. 
We use a network with FC layer, \textbf{6x\_128\_FC} with 4096 units for 100 $\times$ 100 feature input. Note that all though the number of output neurons in this FC layer can be considered to be large (in comparison to classical CNNs for classification), the output dimension is less than the input dimensions which causes some loss in information for generative tasks such as inpainting.

\textbf{Symmetrical skip connections}
For deep network, symmetrical skip connections have shown to perform better for the task of inpainting of images \cite{Mao2016}. The idea is to provide short-cut (addition followed by Relu activation) from the convolutional feature maps to their mirrored transposed-convolution layers in a symmetrical encoding-decoding network. This is particularly helpful with a network with a large depth. In our experiments, we consider a deep network of 12 layers with skip connections \textbf{long\_12x\_SC} and compare with its non connected counter part \textbf{long\_12x}. All the networks are summarized in Figure \ref{table:networks}.

\subsubsection{Training details}
\label{sec:training_details}
3D patches can be straightforwardly extended to images with 1 channel. Instead of pixel value we have height at a perticular 2D bin which can be negative. Depending on the scale the patches are computed, this height can be dependent on the 3D shape it is computed. Therefore, we need to perform dataset normalization before training and testing. 

\textbf{Patch normalization}
We normalize patch set between 0 and 0.83 (= 1/1.2) before training and assign the missing region or hole-masks as 1. This makes the network easily identify the holes during the training - as the training procedure is technically a blind inpainting method. We manually found that, the network has difficulty in reconstructing fine scaled details when this threshold is lowered further (Eg. 1/1.5). The main idea here is to let the network easily identify the missing regions without sacrificing a big part of the input spectrum.

\textbf{Training} We train on the densely overlapped clean patches computed on a set of clean meshes. Square and circular hole-masks of length 0 to 0.8 times the patch length are created randomly on the fly at random locations on the patches with a uniform probability and is passed through the denoising network during the training. The output of the network is matched against the original patches without holes with a soft binary cross entropy loss between 0 and 1. Note that this training scheme is aimed to reconstruct holes less than 0.8 times the patch length. The use of patches of moderate length computed on quad orientations, enables this method to inpaint holes of small to moderate size.

\subsection{Inpainting pipeline}
\label{sec:testing_inpainting}
Testing consists of inpainting holes in a given 3D mesh. This involves patch computation in the noisy mesh, patch inpainting through a generative model, and the reconstruction of the final mesh. 
For a 3D mesh with holes, the regions to be inpainted are completely empty and have no edge connectivity and vertices information. Thus, to establish the final interior mesh connectivity after CNN based patch reconstruction, there has to be a way of inserting vertices and performing triangulation. We use an existing popular \cite{Liepa2003}, for this purpose of hole triangulation to get a connected hole filled mesh based on local geometry.  This hole-triangulated mesh is also used for quad mesh computation on the mesh with holes. This is important as quad mesh computation is affected by the presence of holes. 

\begin{figure}[t]
\centering

\begin{subfigure}{0.45\linewidth}
  \centering
  \includegraphics[width=0.9\linewidth]{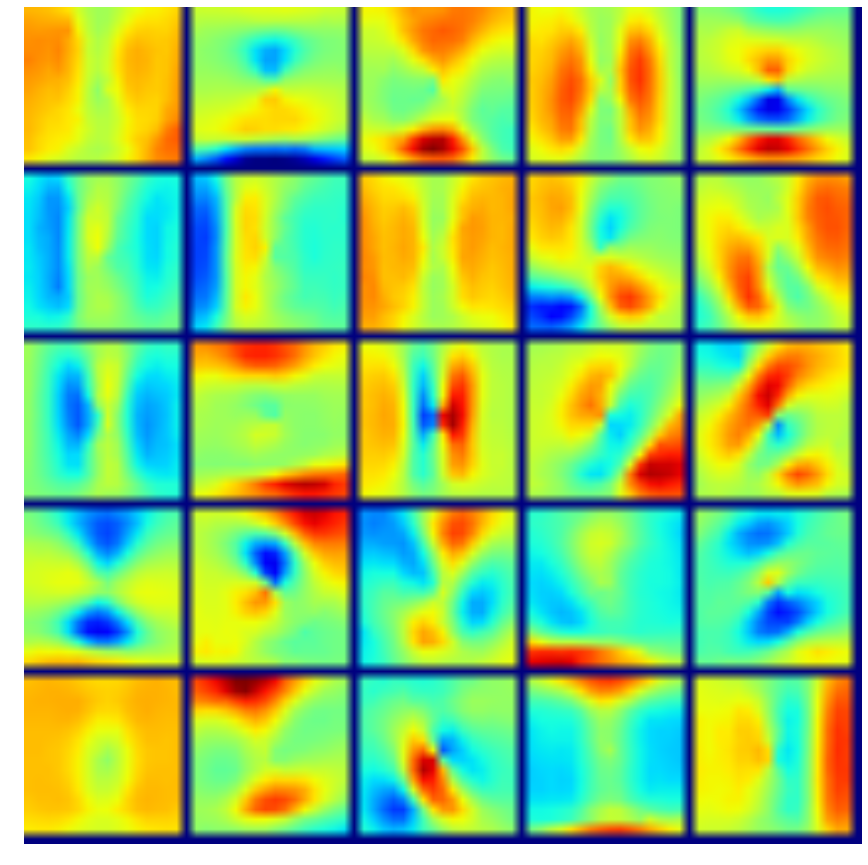}  
\end{subfigure}%
\hfill
\begin{subfigure}{0.5\linewidth}
  \centering
  \includegraphics[width=1\linewidth]{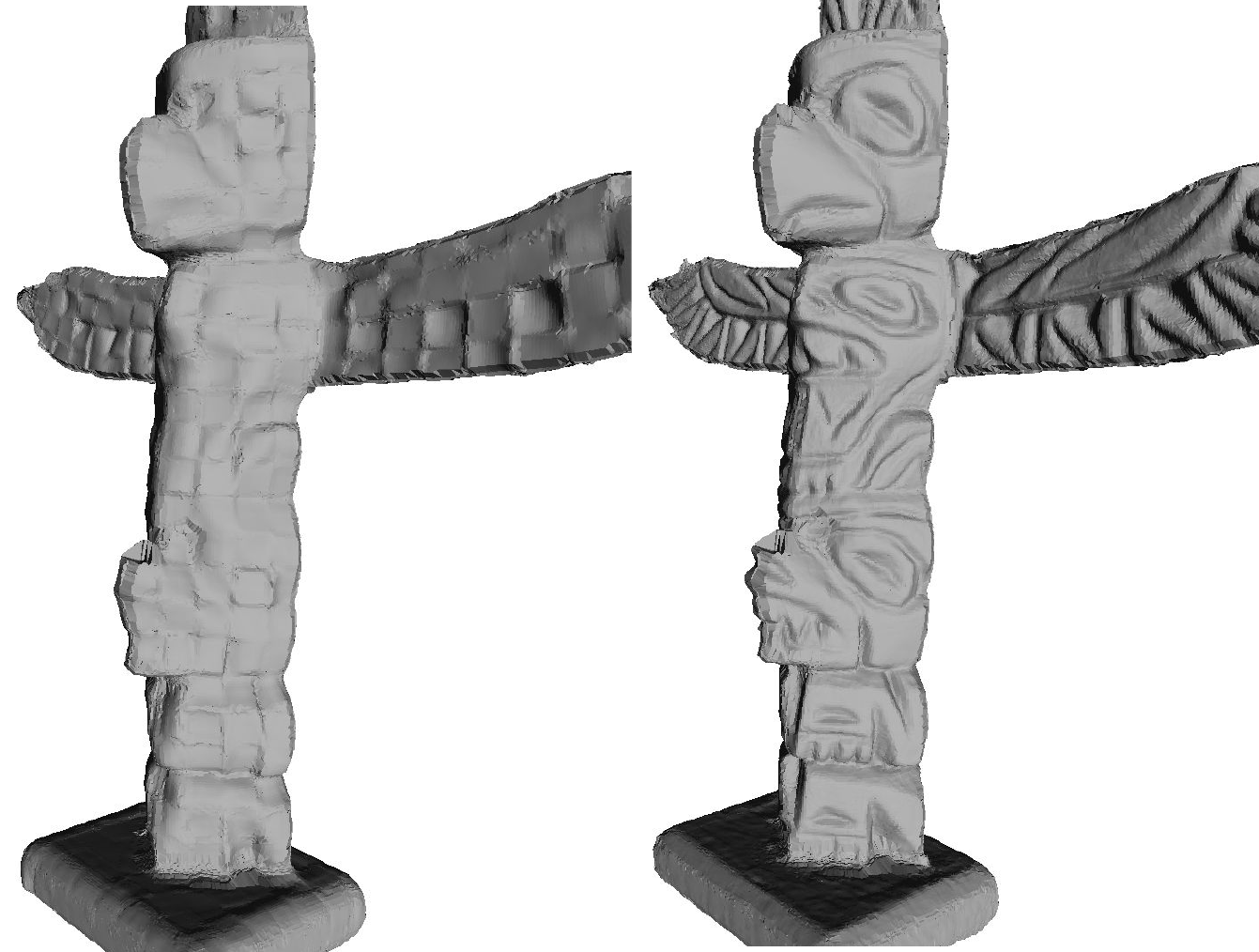}  
\end{subfigure}%
\caption{(Left) Visualization of dictionary atoms learnt from the shape \textit{Totem} ($m = 16 \times 16$). (Right) Reconstruction of the shape \textit{Totem} using local dictionary of size 5 atoms and 100 atoms}
\label{fig:dictionariesvis}
\end{figure}
\section{Experimental Results}
\label{sec:results}
In this section we provide the different experiments performed to evaluate our design choices of both dictionary and CNN based generative models for mesh processing. We first provide the details of the meshes used in our experiments by introducing our dataset in Section \ref{sec:dataset_patches}. We then provide the different parameters used for patch computations followed by the mesh restoration results with dictionary learning (Section \ref{sec:results_patchdict}). We then provide our results of inpainting with CNN based approach and its comparison with our dictionary based approach (Section \ref{sec:conv_results}). As seen both quantitatively and qualitatively, our the CNN based approach provides better results than the dictionary based approach. We finally end up with a section with the generalizing capability of our local patches through global generative models (by both global dictionary and global denoising autoencoder) and discuss the possibility of having a global universal generative model for local 3D patches.

\subsection{Dataset}
\label{sec:dataset_patches}
We considered dataset having 3D shapes of two different types. The first type (\textbf{Type 1}) consists of meshes that are in general, simple in nature without much surface texture.  In descending order of complexity 5 such objects considered are - \emph{Totem, Bunny, Milk-bottle, Fandisk} and \emph{Baseball}. \emph{Totem} is of very high complexity containing a high amount of fine level details whereas \emph{Bunny} and \emph{Fandisk} are some standard graphics models with moderate complexity. In addition, we considered 5 models with high surface texture and details (\textbf{Type 2}) consisting of shoe soles and a human brain specifically to evaluate our hole-filling algorithm - \emph{Supernova, Terrex, Wander, LeatherShoe} and \emph{Brain}. This subset of meshes is also referred as \textit{high texture} dataset in subsequent section. Therefore, we consider in total 10 different meshes for our evaluation (all meshes are shown in the supplementary material). 

Other than the models \textit{Baseball, Fandisk} and \textit{Brain}, all models considered for the experimentation are reconstructed using a vision based reconstruction system - 3Digify \cite{3Digify}. Since this system uses structured light, the output models are quite accurate, but do have inherent noise coming from structured light reconstruction and alignments. Nonetheless, because of its high accuracy, we consider these meshes to be `clean' for computing global patch database. These  models were also reconstructed with varying accuracy by changing the reconstruction environment before considering for the experiments of inpainting. In an extreme case some of these models are reconstructed using Structure From Motion for the purpose of denoising using its `clean' counterpart as described in Section \ref{sec:denoisingres}.

\textbf{Dataset normalization and scale selection}
For normalization, we put each mesh into a unit cube followed by upsampling (by subdivision) or downsampling (by edge collapse) to bring it to a common resolution. After normalization, we obtained the low resolution mesh by applying Laplacian smoothing with 30 iterations. We then performed the automatic quadiangulation procedure of \cite{Ebke2013} on the low resolution mesh, with the targeted number of faces such that, it results an average quad length to be 0.03 for Type 1 dataset and 0.06 for Type 2 dataset (for larger holes); which in turns become the average patch length of our dataset. The procedure of smoothing and generating quad mesh can be supervised manually in order to get better quad mesh for reference frame computation. But, in our varied dataset, the automatic procedure gave us the desired results. 

We then generated 3D patches from each of the clean meshes using the procedure provided in Section \ref{sec:3Dpatches}. We chose the number of bins $N$, to be 16 for Type 1 dataset and 24 for Type 2 dataset; to match the resolution the input mesh. To perform experiment in a common space (global dictionary), we also generated patches with patch dimension of 16 in Type 2 dataset with the loss of some output resolution.

%

\begin{figure}[t]
\centering
\begin{subfigure}{\linewidth}
  \centering
  \includegraphics[width=0.85\linewidth]{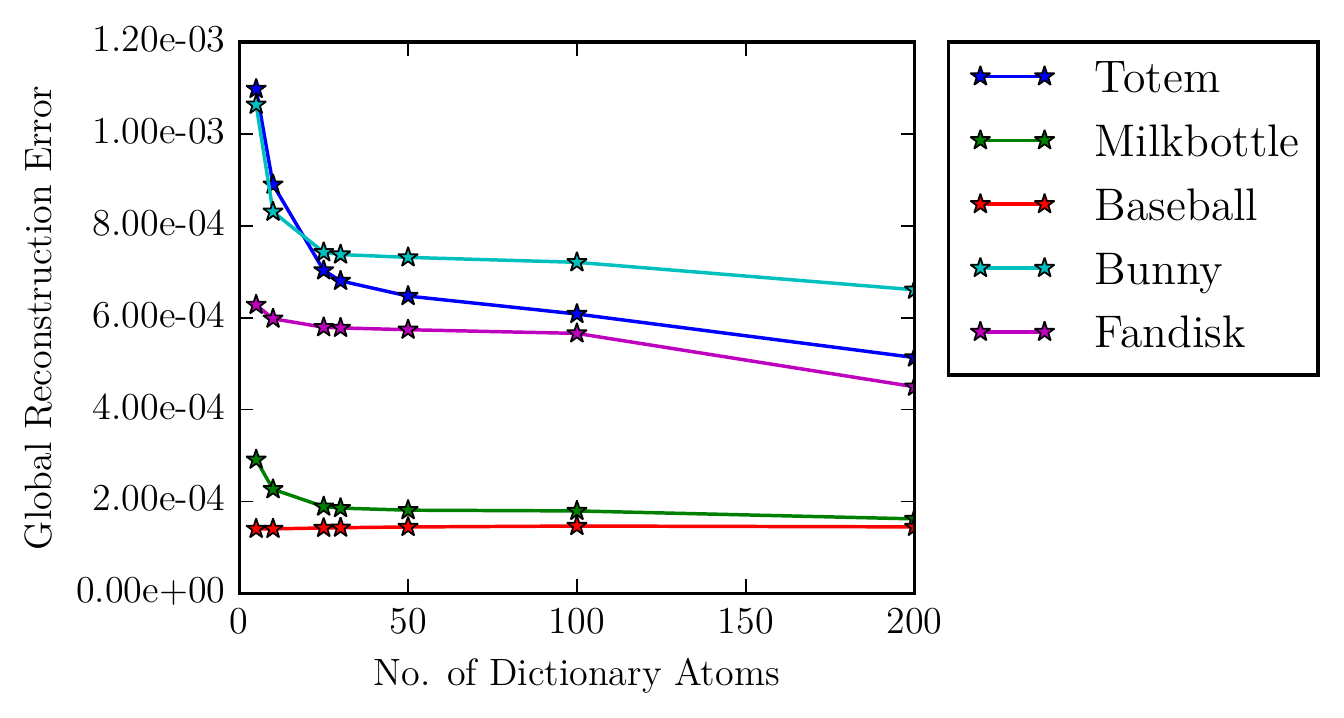}  
\end{subfigure}%
\vspace{-0.4cm}
\caption{Reconstruction error of different shapes with Dictionaries with increasing number of atoms.}
\label{fig:reconstructioncomplexity_quantitative}
\end{figure}

\begin{table}
\small
\centering
\begin{tabular}{cccccc}
\toprule
{} & Mesh &  & Patch & Compr&  \\
{Meshes} & entities & \#patches & entities & factor & PSNR \\
\midrule
Totem      &    450006 &      658 &     12484 & 36.0 & 56.6 \\
Milkbottle &    441591 &      758 &     14420 & 30.6 & 72.3 \\
Baseball   &    415446 &      787 &     14974 & 27.7 & 75.6 \\
Bunny      &    501144 &      844 &     16030 & 31.3 & 60.6 \\
Fandisk    &     65049 &      874 &     16642 &  3.9 & 62.1 \\
\bottomrule
\end{tabular}
\caption{Results for compression in terms of number of entities  with a representation with global dictionary of 100 atoms. Mesh entities consists of the number of entities for representing the mesh which is: 3 $\times$ \#Faces and \#Vertices. Patch entities consists of the total number of sparce dictionary coefficients (20 per patch) used to represent the mesh plus the entities in the quad mesh. Compr factor is the compression factor between the two representation. PSNR is Peek Signal to Noise Ratio where the bounding box diameter of the mesh is considered as the peek signal following \cite{Praun2003}}.

\label{table:compression}
\end{table}
\subsection{Evaluating 3D Patch Dictionaries}
\label{sec:results_patchdict}
\subsubsection{Dictionary Learning and Mesh Reconstruction}

\textbf{Dictionary Learning}
We learn the local dictionary for each shape with varying numbers of dictionary atoms with the aim to reconstruct the shape with varying details. Atoms of one such learned dictionary is shown in Figure \ref{fig:dictionariesvis} (Left). Observe the `stripe like' structures the dictionary of \textit{Totem} in accordance to the fact that the \textit{Totem} has more line like geometric textures. 

\textbf{Reconstruction of shapes}
We then perform reconstruction of the original shape using the local dictionaries with different number of atoms (Section \ref{sec:shapeencoding}). 
Figure \ref{fig:dictionariesvis} (Right) shows qualitatively the difference in output shape when reconstructed with dictionary with 5 and 100 atoms. 
Figure \ref{fig:reconstructioncomplexity_quantitative} shows the plot between the \textit{Global Reconstruction Error} - the mean Point to Mesh distance of the vertices of the reconstructed mesh and the reference mesh - and the number of atoms in the learned dictionary for our Type 1 dataset. We note that the reconstruction error saturates after a certain number of atoms (50 for all).

\textbf{Potential for Compression} The reconstruction error is low after a certain number of atoms in the learned dictionary, even when global dictionary is used for reconstructing all the shapes (more on shape independence in Section \ref{sec:generalization}). Thus, only the sparse coefficients and the connectivity information needs to be stored for a representation of a mesh using a common global dictionary, and can be used as a means of mesh compression. Table \ref{table:compression} shows the results of information compression on Type 1 dataset. 

\begin{figure}
\centering
\begin{subfigure}{1\linewidth}
  \centering
  \includegraphics[width=\linewidth]{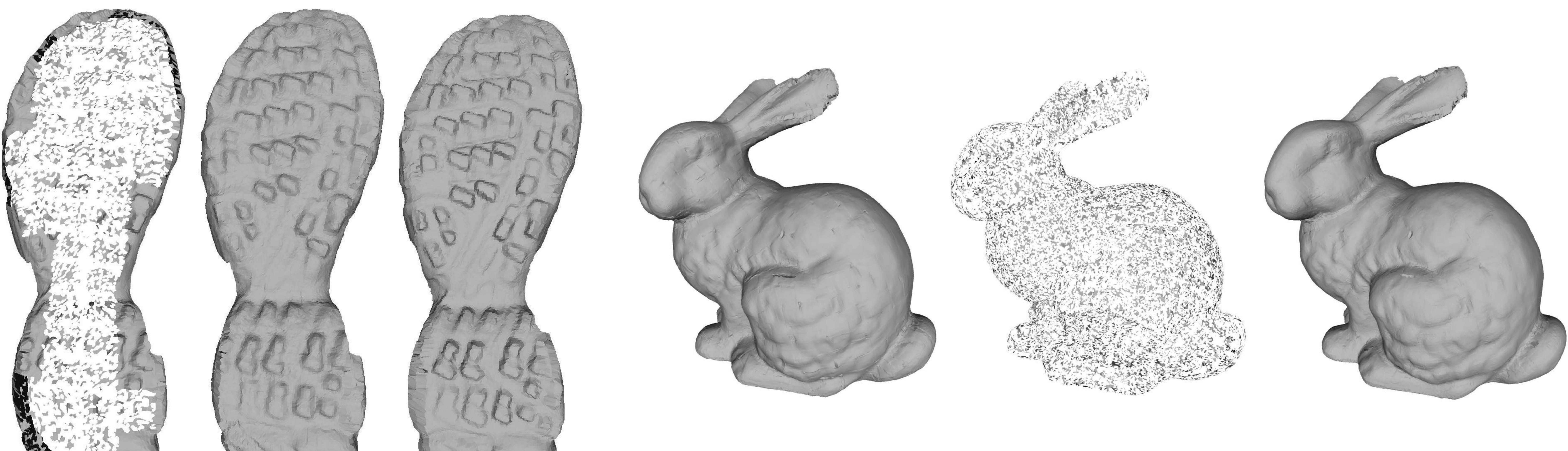}  
\end{subfigure}
\caption{Inpainting of the models with 50\% missing vertices (Left - noisy mesh, Middle - inpainted mesh, Right - ground truth) of \textit{Terrex} and \textit{Bunny}, using the local dictionary. Here we use the quad mesh provided at the testing time.}
\label{fig:missingvertices}

\end{figure}

\begin{table}
\centering
\small
\begin{tabular}{l|cc|cc}
\hline
{Missing Ratio} &       \multicolumn{2}{c|}{0.2} &       \multicolumn{2}{c}{0.5}   \\
{} & ours & \cite{Zhong2016} & ours & \cite{Zhong2016} \\
\hline

bunny  & \textbf{1.11e-3} &1.90e-2 &\textbf{1.62e-3} & 2.20e-2   \\
fandisk &  \textbf{1.32e-3} &8.30e-3 & \textbf{1.34e-3} &1.20e-2\\
\hline
\end{tabular}
\caption{RMS Inpainting error of missing vertices from our method using local dictionary and its comparison to \cite{Zhong2016}}.
\label{table:zhongcomp}

\end{table}

\begin{figure}[t]
\centering
\begin{subfigure}{1\linewidth}
  \centering
  \includegraphics[width=.75\linewidth]{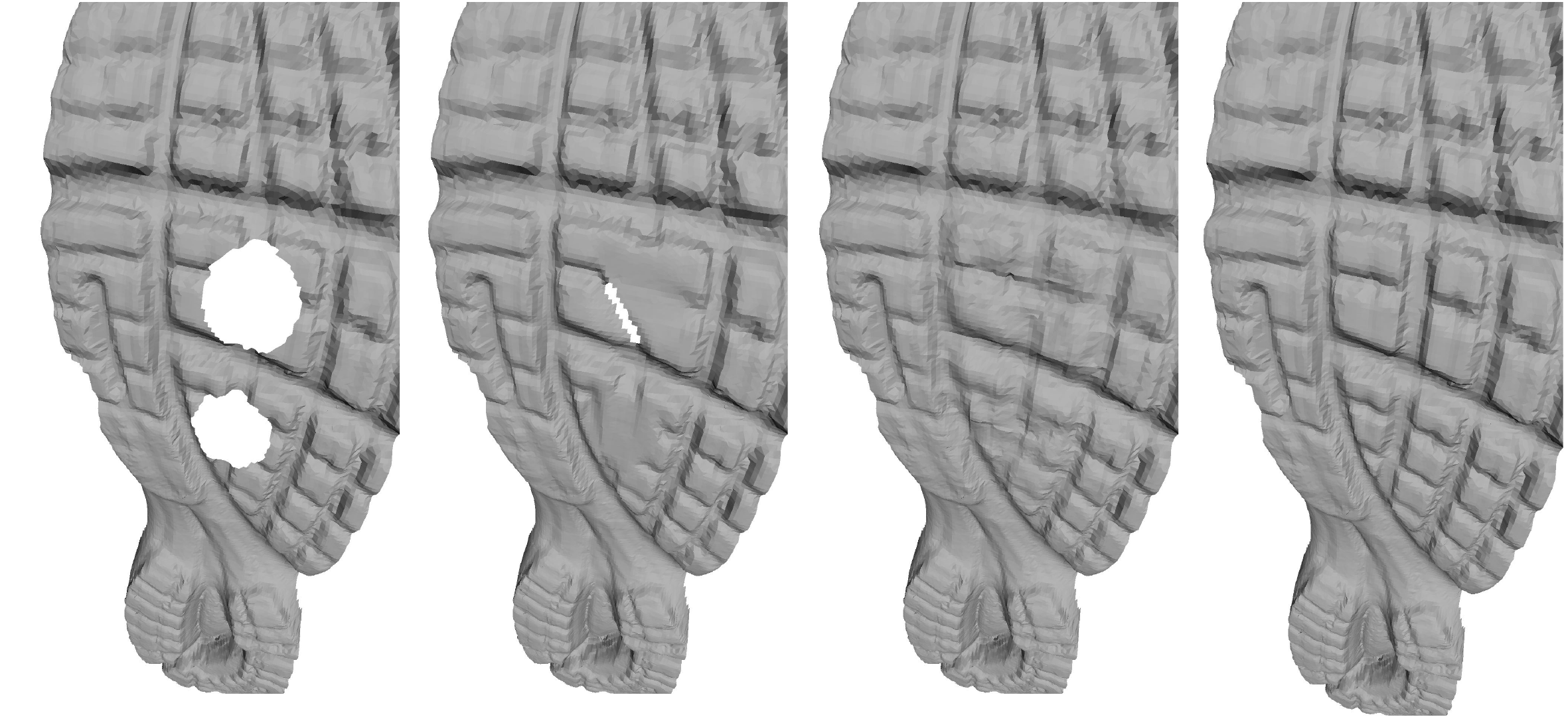}  
\end{subfigure}%

\begin{subfigure}{1\linewidth}
  \centering
  \includegraphics[width=0.75\linewidth]{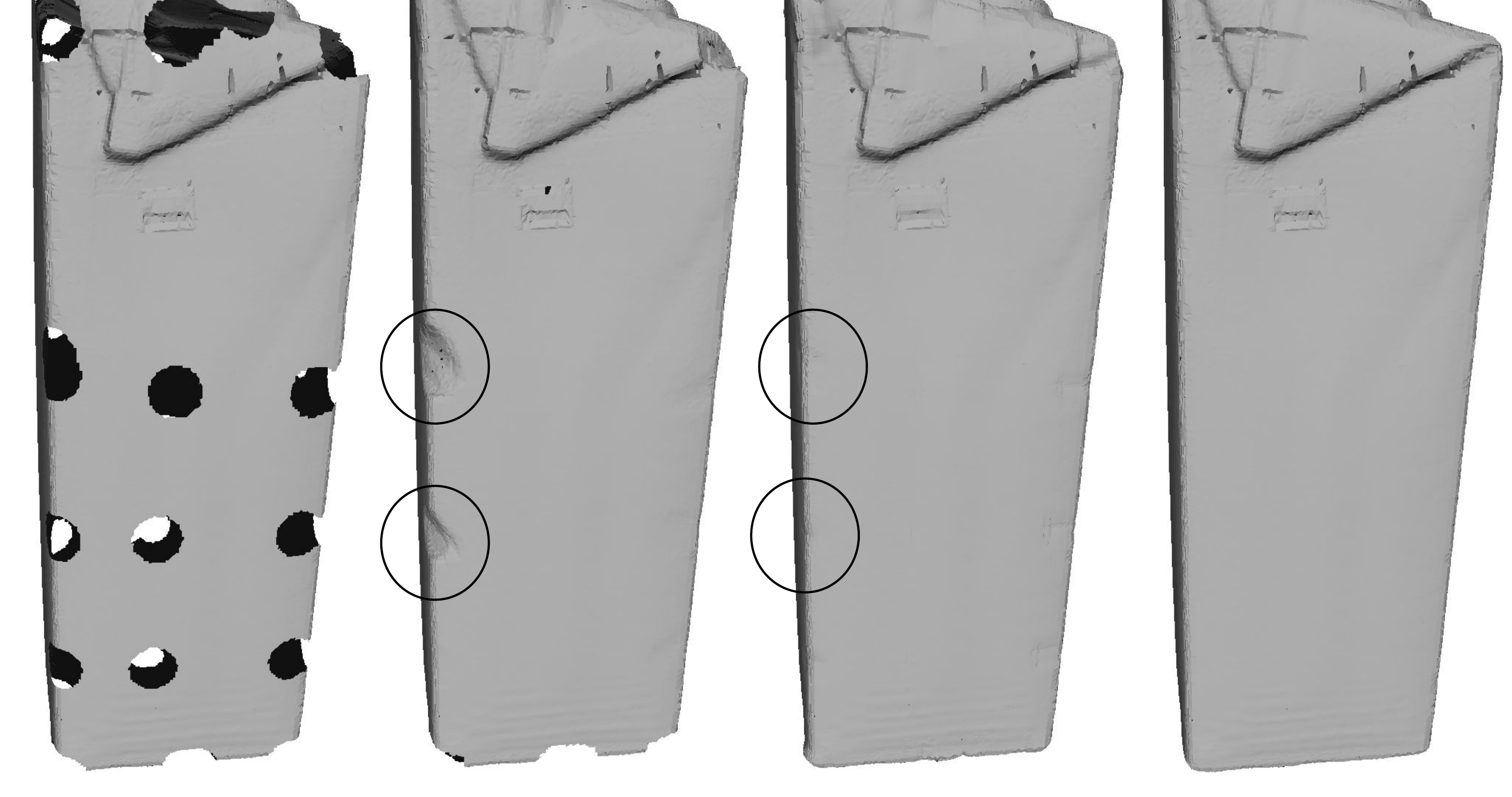}  
\end{subfigure}%
\caption{Qualitative analysis of the inpainting algorithm of \textit{Supernova} and \textit{Milk-bottle}. From left to right - mesh with holes, hole filling with  \cite{Liepa2003}, our results from global dictionary and ground truth mesh. Detailed visualization of the results of other meshes are presented in the provided supplementary material.}
\label{fig:inpaintqualitative}

\end{figure}

\subsubsection{Surface Inpainting}

\textbf{Recovering missing geometry}
To evaluate our algorithm for geometry recovery, we randomly label certain percentage of vertices in the mesh as missing. The reconstructed vertices are then compared with the original ones. The visualization of our result is in Figure \ref{fig:missingvertices}. Zoomed view highlighting the details captured as well as the results from other objects are provided in the supplementary material. We compare our results with \cite{Zhong2016} which performs similar task of estimating missing vertices, with the publically available meshes \textit{Bunny} and \textit{Fandisk}, and provide the recovery error measured as the Root Mean Square Error (RMSE) of the missing coordinates in Table \ref{table:zhongcomp}. Because of the unavailability of the other two meshes used in \cite{Zhong2016}, we limited to these aforementioned meshes for comparison. As seen in the table, we improve over them by a large margin.

This experiment also covers the case when the coarse mesh of the noisy data is provided to us which we can directly use for computing quad mesh and infer the final mesh connectivity (Section \ref{sec:connectedmeshrec}). This is true for the application of recovering damaged part. If the coarse mesh is not provided, we can easily perform poisson surface reconstruction using the non-missing vertices followed by Laplacian smoothing to get our low resolution mesh for quadriangulation. Since, the low resolution mesh is needed just for the shape outline without any details, poisson surface reconstruction does sufficiently well even when 70\% of the vertices are missing in our meshes.

\textbf{Hole filling}
We systematically punched holes of different size (limiting to the patch length) uniform distance apart in the models of our dataset to create noisy test dataset. We follow the procedure in Section \ref{sec:testing_inpainting} in this noisy dataset and report our inpainting results in Table \ref{table:inpaintingall}. Here we use mean of the Cloud-to-Mesh error of the inpainted vertices as our error metrics.  Please note that the noisy patches are independently generated on its own quad mesh. No information about the reference frames from the training data is used for patch computation of the noisy data. Also, note that this logically covers the inpainting of the missing geometry of a scan due to occlusions. We use both local and global dictionaries for filling in the missing information and found our results to be quite similar to each other.   

For baseline comparison we computed the error from the popular filling algorithm of \cite{Liepa2003} available in MeshLab\cite{Cignoni2008}. Here the comparison is to point out the improvement achieved using a data driven approach over geometry. We could not compare our results with \cite{Zhong2016} because of the lack of systematic evaluation of hole-filling in their paper. As it is seen, our method is clearly better compared to the \cite{Liepa2003} quantitatively and qualitatively (Figure \ref{fig:inpaintqualitative}). The focus of our evaluation here is on the Type 2 dataset - which captures complex textures. In this particular dataset we also performed the hole filling procedure using self-similarity, where we learn a dictionary from the patches computed on the noisy mesh having holes, and use it to reconstruct the missing data. The results obtained is very similar to the use of local or global dictionary (Table \ref{table:inpaintselfsimilar}).

\begin{figure}
\centering
\begin{subfigure}{0.65\linewidth}
  \centering
  \includegraphics[width=1\linewidth]{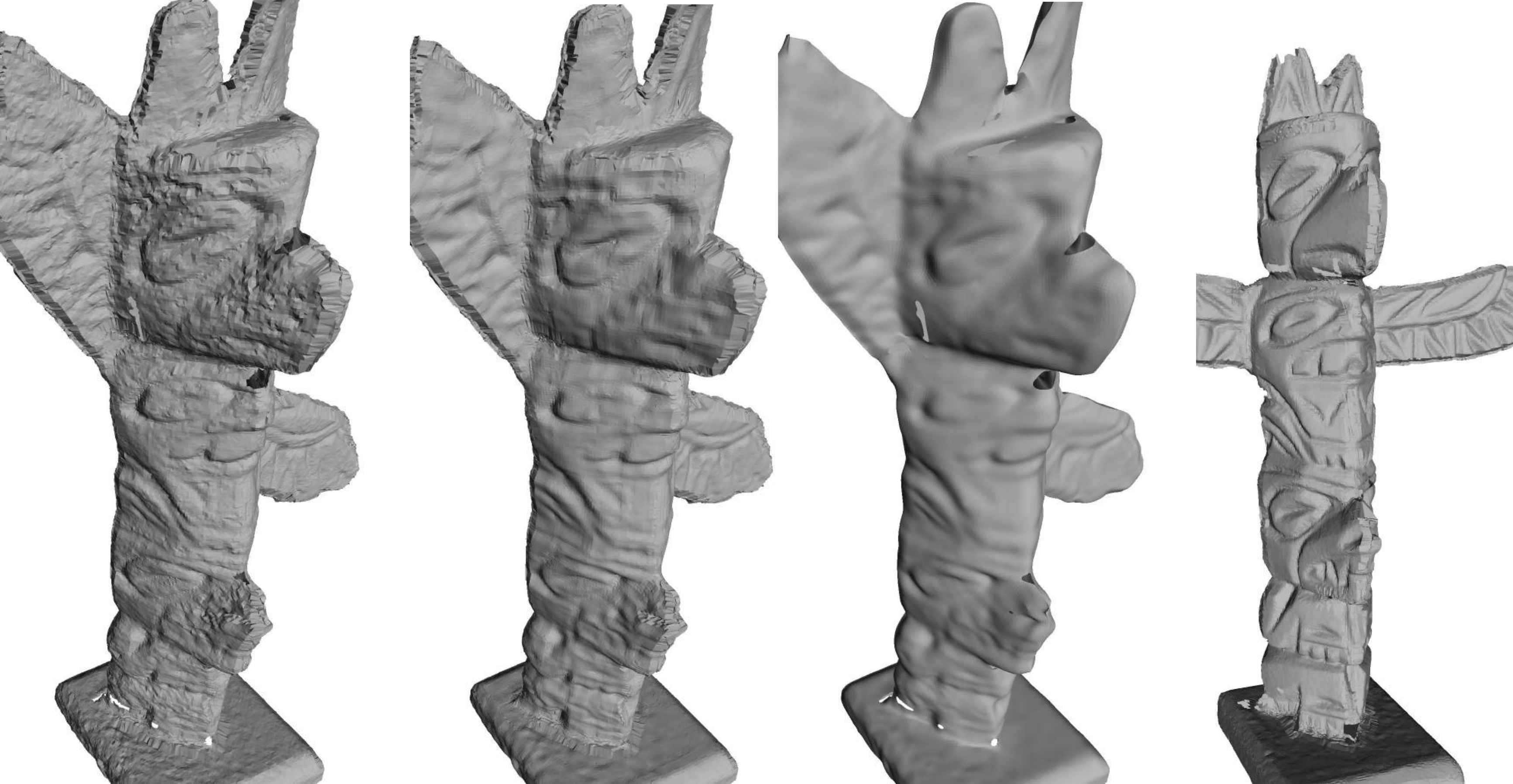}  
\end{subfigure}\begin{subfigure}{0.35\linewidth}\centering\includegraphics[width=0.6\linewidth]{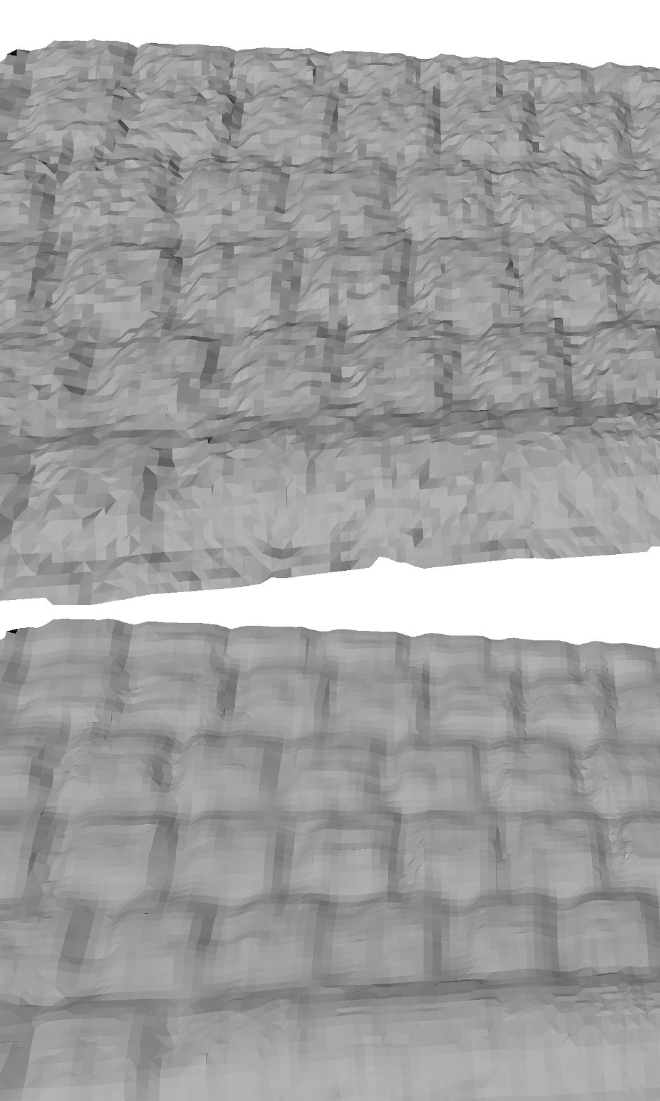}  
\end{subfigure}%
\caption{Denoising meshes using a clean patch dictionary of a similar object. (Left) Results on \textit{Totem} (from left to right) - noisy reconstruction from SFM, our denoising using patch dictionary from a clean reconstruction, denoising by Laplacian smoothing \cite{Sorkine2004}, the high quality clean mesh with different global configuration. (Right) Result for the mesh \textit{Keyboard} with the same experiment. Zoomed versions of similar results are provided in the supplementary material.}
\label{fig:denoising_qualitative}
\end{figure}

\begin{table}[t]
\centering
\small
\begin{tabular}{lrrr}
\toprule
{} &   \cite{Liepa2003} &     Our - Local &    Our - Global \\
\midrule
Supernova &  0.001646 &  \textbf{0.000499} &  0.000524 \\
Terrex     &  0.001258 &  0.000595 &  \textbf{0.000575} \\
Wander     &  0.002214 &  0.000948 &  \textbf{0.000901} \\
LeatherShoe &  0.000854 &  0.000569 &  \textbf{0.000532} \\
Brain      &  0.002273 &  0.000646 &  \textbf{0.000587} \\ \hline
Milk-bottle &  0.000327 &  0.000126 &  \textbf{0.000123} \\
Baseball   &  0.000158 &  \textbf{0.000138} &  0.000168 \\
Totem      &  0.001065 &  0.001065 &  \textbf{0.001052} \\
Bunny      &  \textbf{0.000551} &  0.000576 &  0.000569 \\
Fandisk    &  0.001667 &  0.000654 &  \textbf{0.000634} \\
\bottomrule
\end{tabular}
\caption{Mean inpainting error for our dataset of hole size 0.015, 0.025 and 0.035 for the dataset Type 2 (top block of the table) and 0.01 and 0.02 for Dataset Type 1 (bottom block of the table). \textit{Local} uses the local dictionary learned from the clean mesh of the corresponding shape and \textit{Global} uses a global dictionary learned from the entire dataset.}
\label{table:inpaintingall}

\end{table}

\begin{table}[th]
\centering
\small
\begin{tabular}{lrr}
\toprule
{} &   \cite{Liepa2003} &  Self-Similar \\
\midrule
Supernova &  0.001162 &     \textbf{0.000401} \\
Terrex     &  0.000900 &    \textbf{0.000585} \\
Wander     &  0.001373 &     \textbf{0.000959} \\
LeatherShoe &  0.000596 &     \textbf{0.000544} \\
Brain      &  0.001704 &     \textbf{0.000614} \\
\bottomrule
\end{tabular}
\caption{Mean inpainting error comparison with self similar dictionary with 100 atoms. Hole size considered is 0.035}
\label{table:inpaintselfsimilar}
\end{table}

\begin{figure*}[th]
\centering
\begin{subfigure}[b]{0.33\linewidth}
  \centering
  \includegraphics[width=0.99\linewidth]{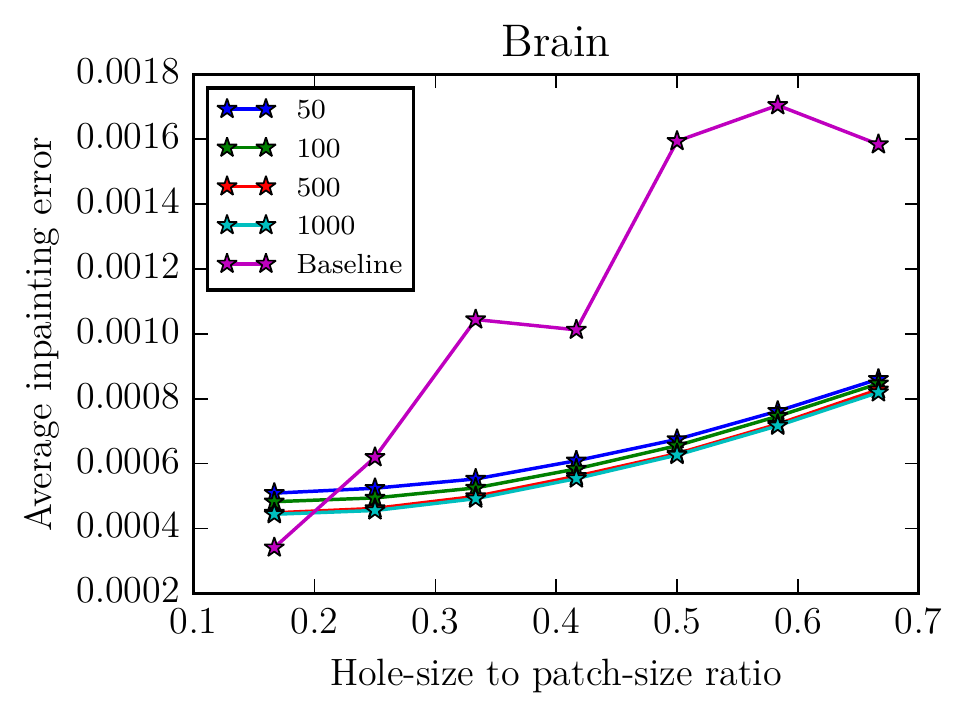} 
  \vspace{-0.25cm}
  \caption{} 
  \label{fig:brain_holewise_algo}
\end{subfigure}\begin{subfigure}[b]{0.33\linewidth}
  \centering
  \includegraphics[width=0.99\linewidth]{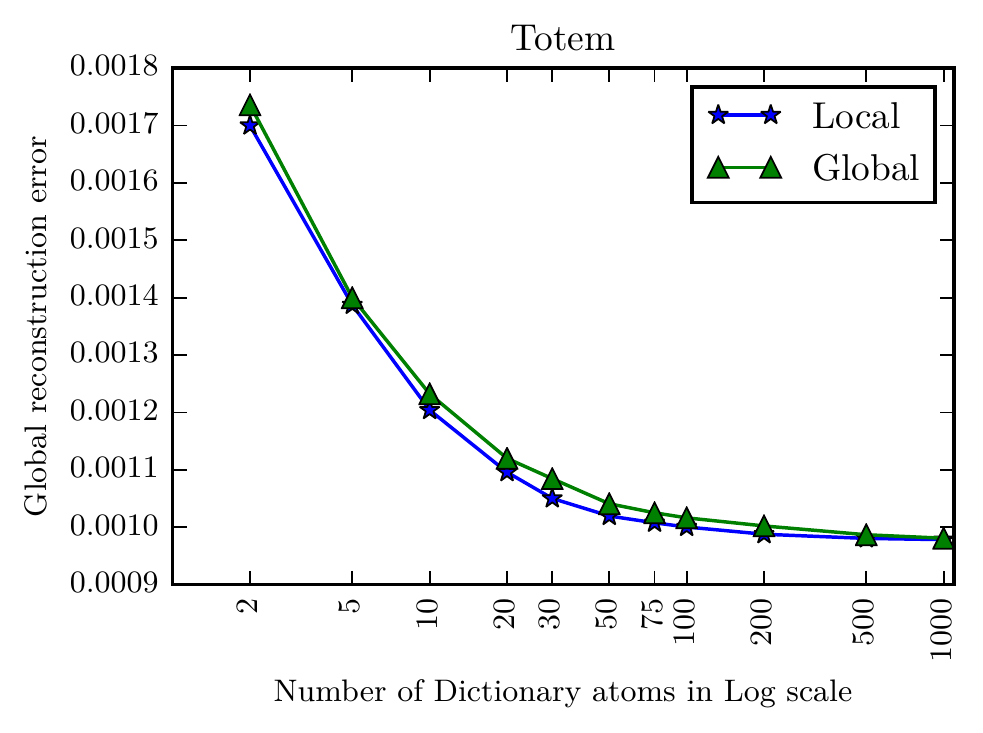}    
    \vspace{-0.25cm}
  \caption{} 
  \label{fig:recerror_global_local}
\end{subfigure}\begin{subfigure}[b]{0.33\linewidth}
  \centering
  \includegraphics[width=0.99\linewidth]{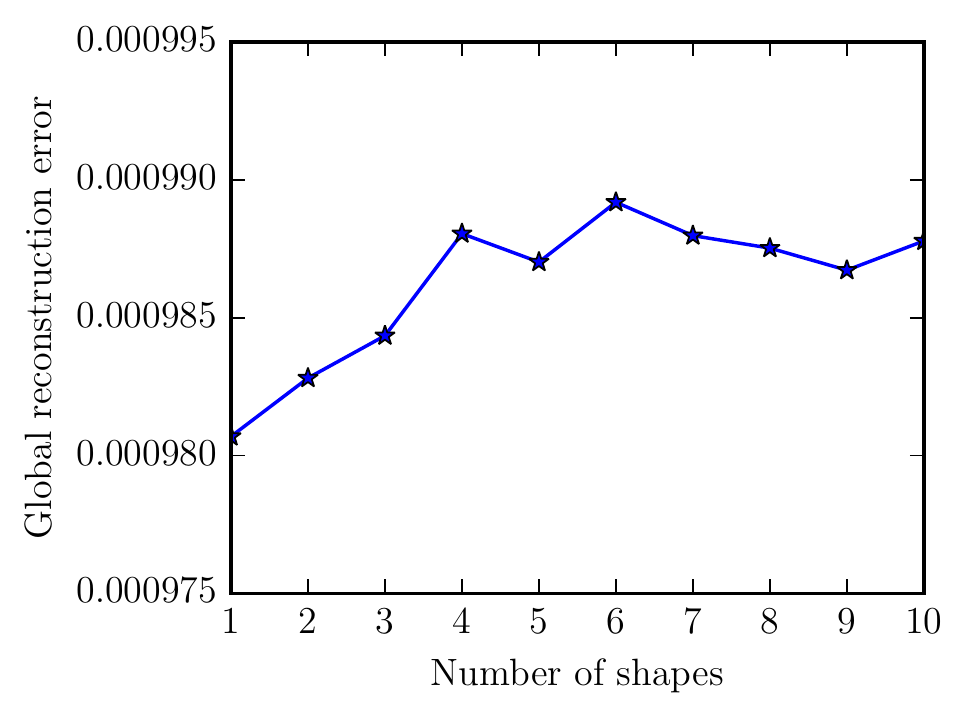} 
    \vspace{-0.25cm}
  \caption{}  
    \label{fig:reconstructioncomplexity}
\end{subfigure}%
\vspace{-0.4cm}
\caption{(a) Inpainting error vs Hole-size to patch-size ratio for \textit{Brain} inpainted using the global dictionary. The patchsize here is 0.062 (patch radius $\approx$ 0.044). Plots of other shapes are in provided in the supplementary material (b) Comparison of the reconstruction error of \textit{Totem} using local and global dictionaries with different number of atoms. For better visualization the X axis provided in logarithmic scale. (c) Reconstruction error of \textit{Totem} with global dictionaries (with 500 atoms) having patches from different number of shapes.}
\end{figure*}

With smaller holes, the method of \cite{Liepa2003} performs as good as our algorithm, as shape information is not present in such a small scale. The performance of our algorithms becomes noticeably better as the hole size increases as described by Figure \ref{fig:brain_holewise_algo}. This shows the advantage of our method for moderately sized holes. 

\textbf{Improving quality of noisy reconstruction}
\label{sec:denoisingres}
Our algorithm for inpainting can be easily extended for the purpose of denoising. We can use the dictionary learned on the patches from a clean or high quality reconstruction of an object to improve the quality of its low quality reconstruction. Here we approximate the noisy patch with its closest linear combination in the Dictionary following Equation \ref{eq:sparsity}. Because of the fact that our patches are local, the low quality reconstruction need not be globally similar to the clean shape. This is depicted by the Figure \ref{fig:denoising_qualitative} (Left) where a different configuration of the model \textit{Totem} (with the wings turned compared to the horizontal position in its clean counterpart) reconstructed with structure-from-motion with noisy bumps has been denoised by using the patch dictionary learnt on its clean version reconstructed by Structured Light. A similar result on Keyboard is shown in Figure \ref{fig:denoising_qualitative} (Right).


\subsection{Evaluating Denoising Autoencoders}
\label{sec:conv_results}

We use the same dataset mentioned in the beginning of this section for evaluating Convolutional Denoising Autoencoders and put more emphasis to the \textit{high texture dataset}. We compute another set of patches with resolution 100 $\times$ 100 (in addition to computing patches with resolution 24 $\times$ 24 as presented in Section \ref{sec:dataset_patches}) for performing fine level analysis of patch reconstruction w.r.t. the network complexity.


\begin{figure*}
\centering
\begin{subfigure}{\linewidth}
  \centering
  \includegraphics[width=0.85\linewidth]{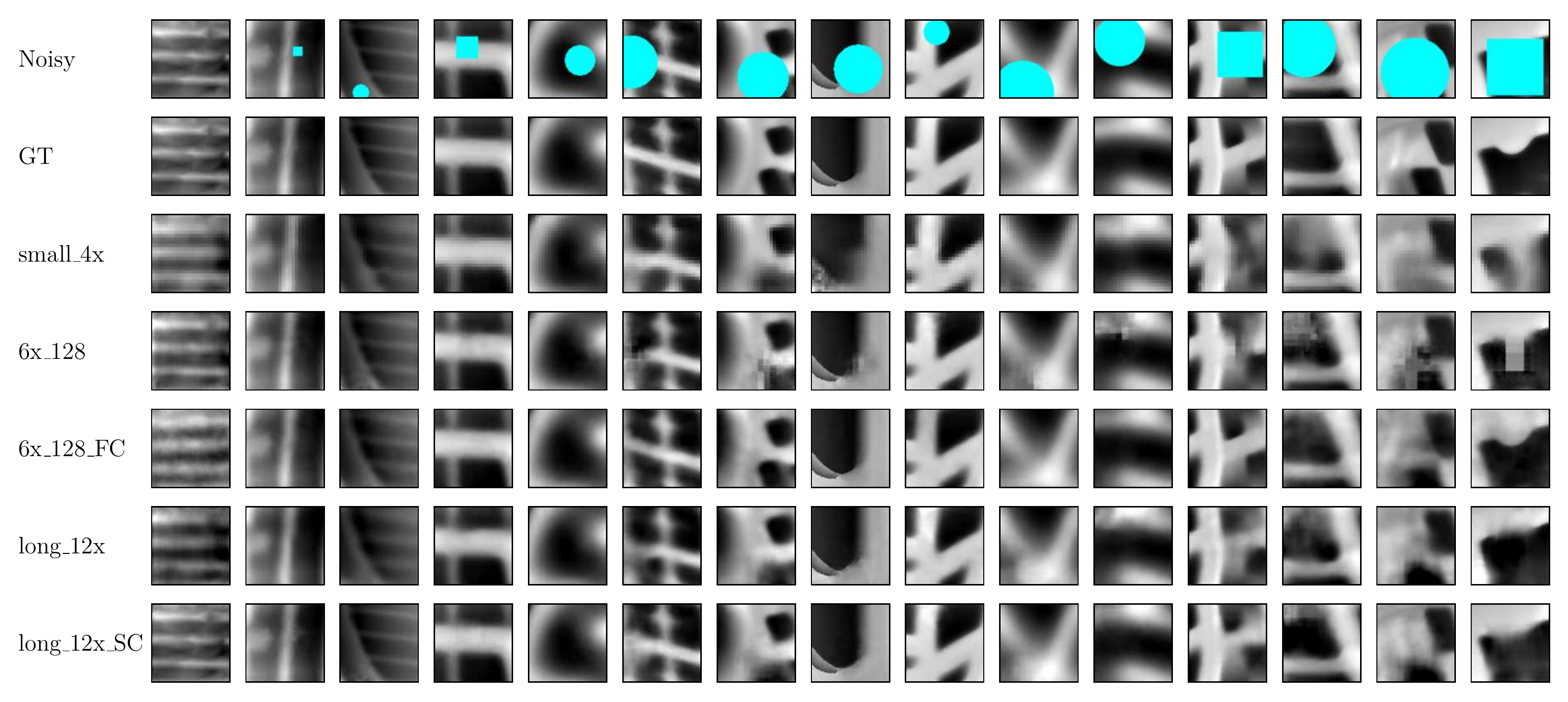}  
\end{subfigure}%
\caption{Qualitative result of our inpainting method with different patches of dimension 100 $\times$ 100 (24 $\times$ 24 for \textit{small\_4x}) with global networks. Patches are taken at random from the test set of meshes of shoe soles and brain, and random masks of variable size, shown in cyan (light blue), are chosen for the task of inpainting. Results of the inpainted patches with different network architectures are shown in the bottom rows. }
\label{fig:qualitative_test}
\end{figure*}

\begin{table*}
\centering
\small
\begin{tabular}{l|rr|rrrrrrr}
\toprule
Meshes &   \cite{Liepa2003} &   Global Dict    &    \textbf{small\_4x}    & \textbf{multi\_6x} & \textbf{6x\_128} & \textbf{6x\_128\_FC} & \textbf{l\_12x} & \textbf{l\_12x\_SC} \\
\midrule
Supernova &  0.001646  & 0.000524  &  0.000427 & 0.000175 &  0.000173 &  0.000291 &  0.000185 &  0.000162 \\
Terrex     &  0.001258    &  0.000575 &  0.000591 & 0.000373 &  0.000371 &  0.000488 &  0.000395 &  0.000369\\
Wander     &  0.002214  &  0.000901 &  0.000894 & 0.000631 &  0.000628 &  0.001033 &  0.000694 &  0.000616 \\
LeatherShoe &  0.000854  &  0.000532 &  0.000570 & 0.000421 &  0.000412 &  0.000525 &  0.000451 &  0.000407 \\
Brain      &  0.002273  &  0.000587 &  0.000436 & 0.000166 &  0.000171 &  0.000756 &  0.000299 &  0.000165 \\
\bottomrule
\end{tabular}
\caption{Mean inpainting error for our dataset of shoe soles of hole size 0.015, 0.025 and 0.035 with a single CNN of different architecture and its comparison to the global dictionary based method. As expected, the error decreases with the increase in the complexity (network length, skip connections, etc).}

\label{table:inpaintingall}
\end{table*}


\textbf{Training and Testing} We train different CNNs from the clean meshes as described in the following sections. For testing or hole filling, we systematically punched holes of different size (limiting to the patch length) uniform distance apart in the models of our dataset to create noisy test dataset. The holes are triangulated to get connectivity as described in the Section \ref{sec:testing_inpainting}. Finally, noisy patches are generated on a different set of quad-mesh (Reference frames) computed on the hole triangulated mesh, so that we use a different set of patches during the testing. More on the generalising capability of the CNNs are discussed in the Section \ref{sec:generalization}.


\begin{figure*}
\centering
\small
\begin{tabular}{|cccccc|}

Holes &  GT & \cite{Liepa2003} &     Global Dict &  small\_4x & long\_12x\_SC \\
\includegraphics[width=\tablequalwidthWACV\linewidth]{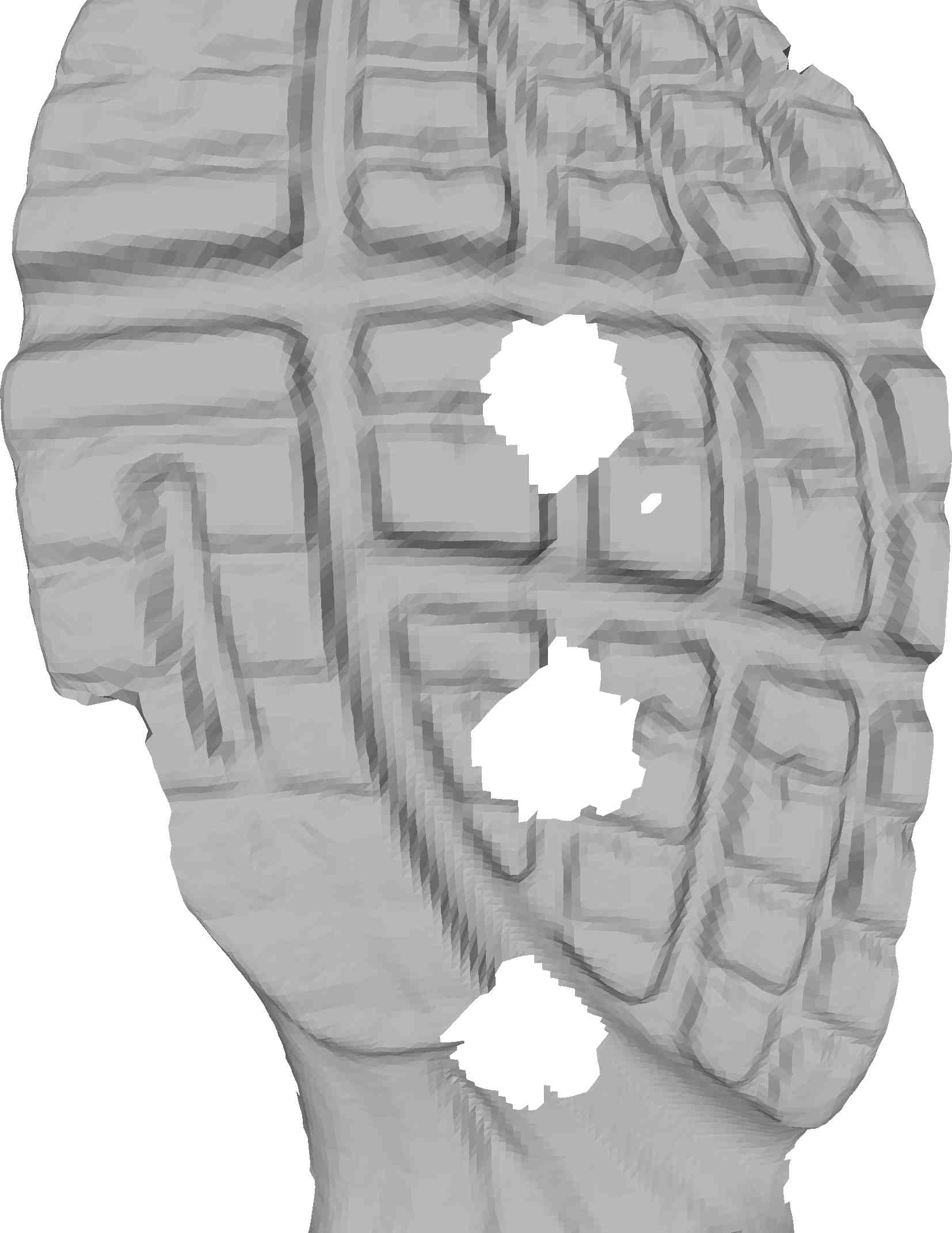}&
\includegraphics[width=\tablequalwidthWACV\linewidth]{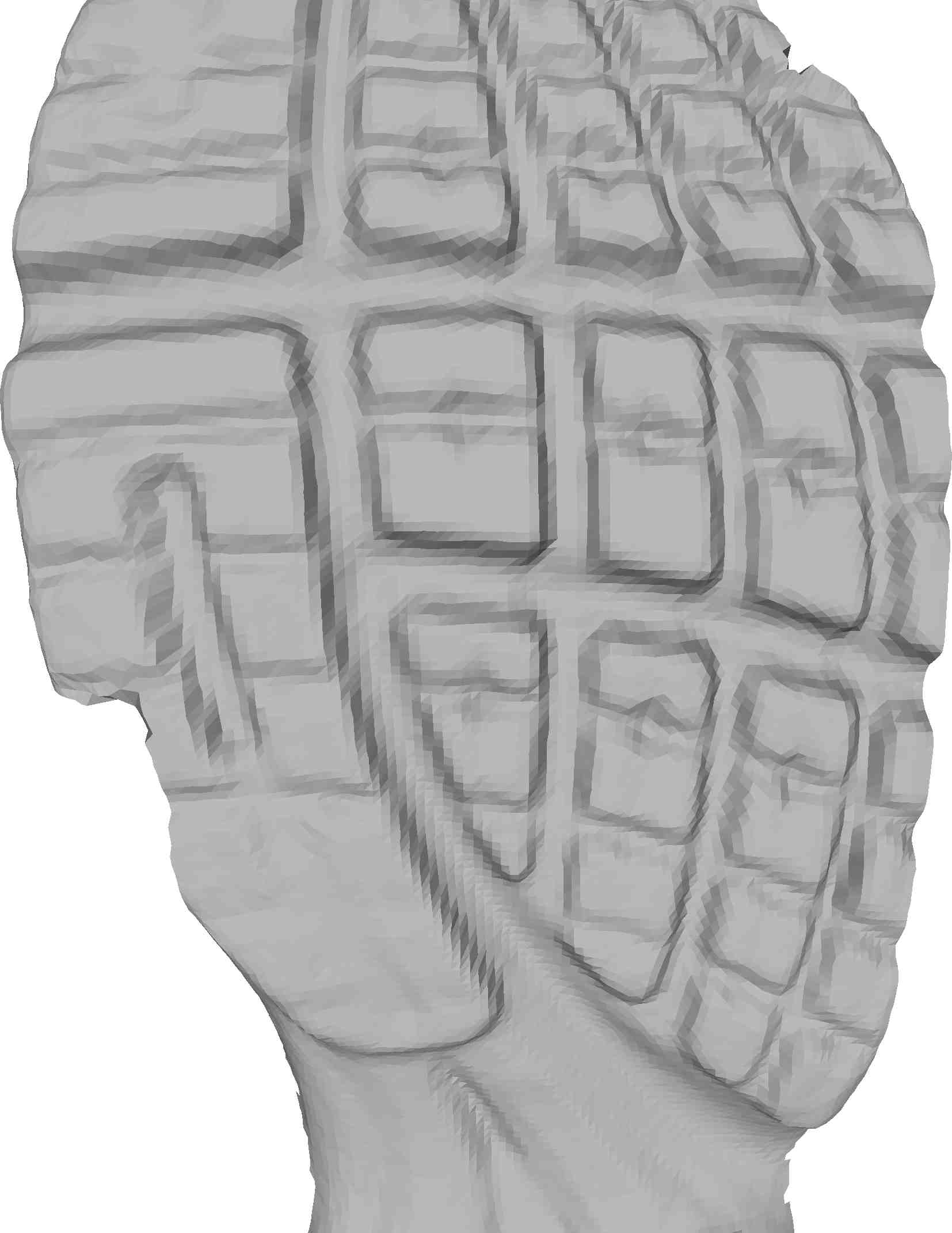}&
\includegraphics[width=\tablequalwidthWACV\linewidth]{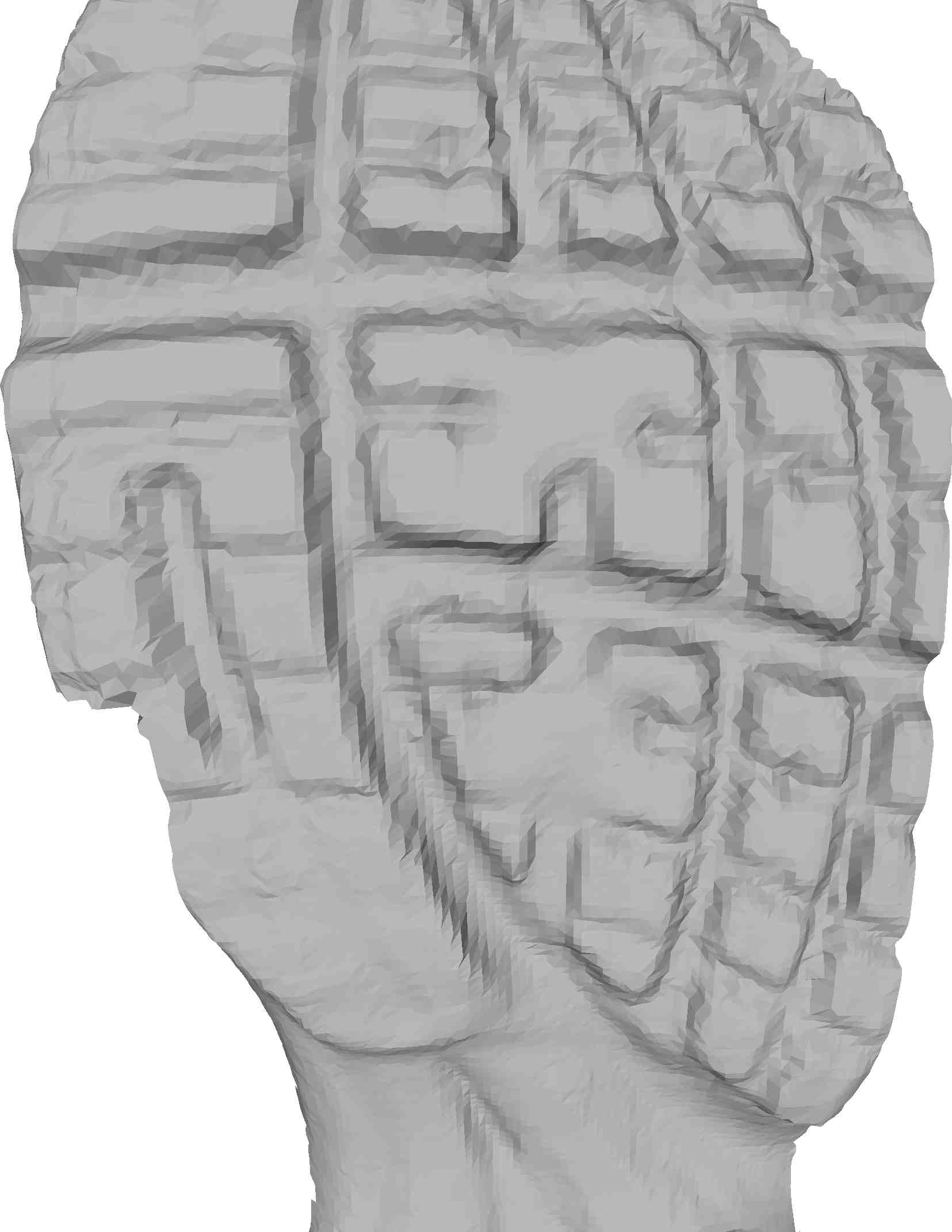}&
\includegraphics[width=\tablequalwidthWACV\linewidth]{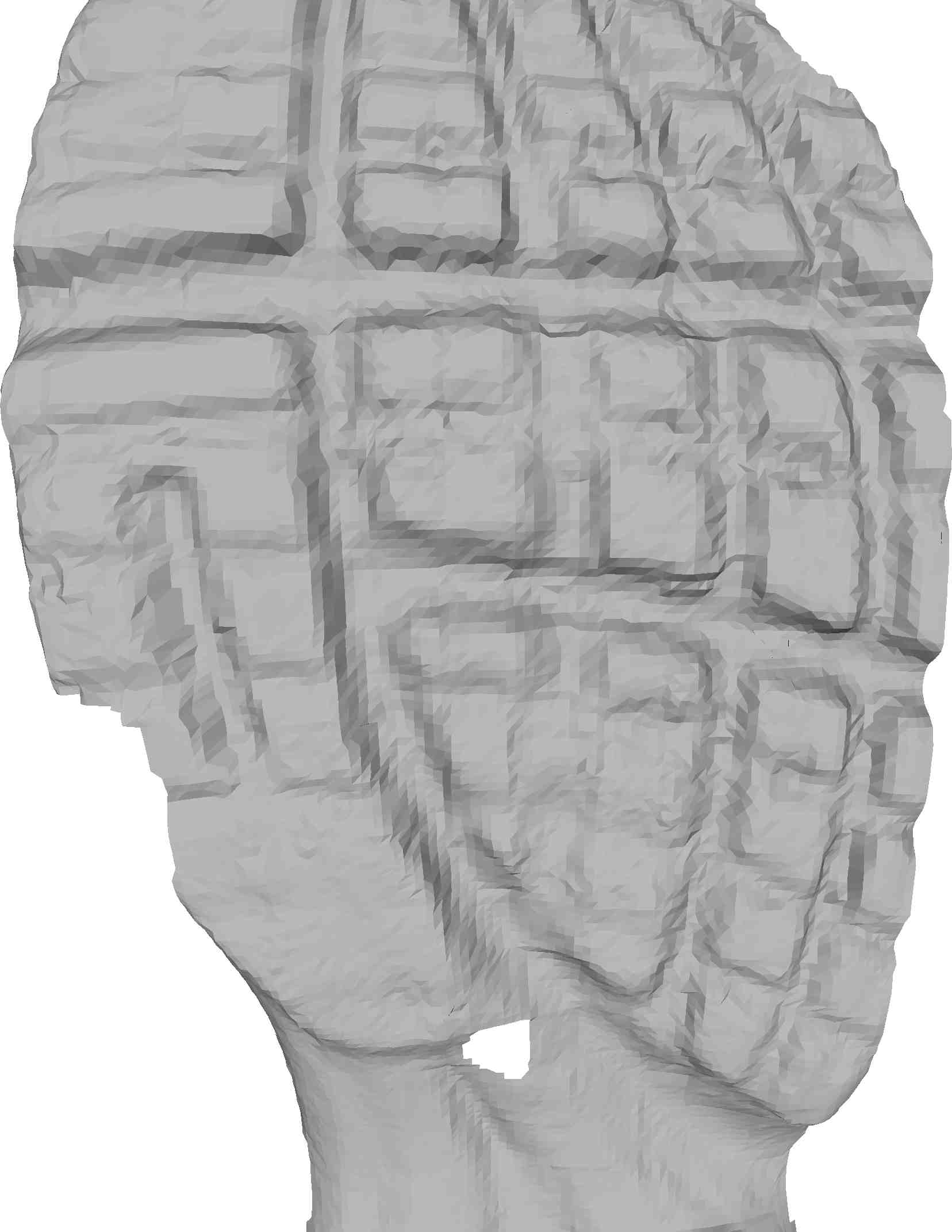}&
\includegraphics[width=\tablequalwidthWACV\linewidth]{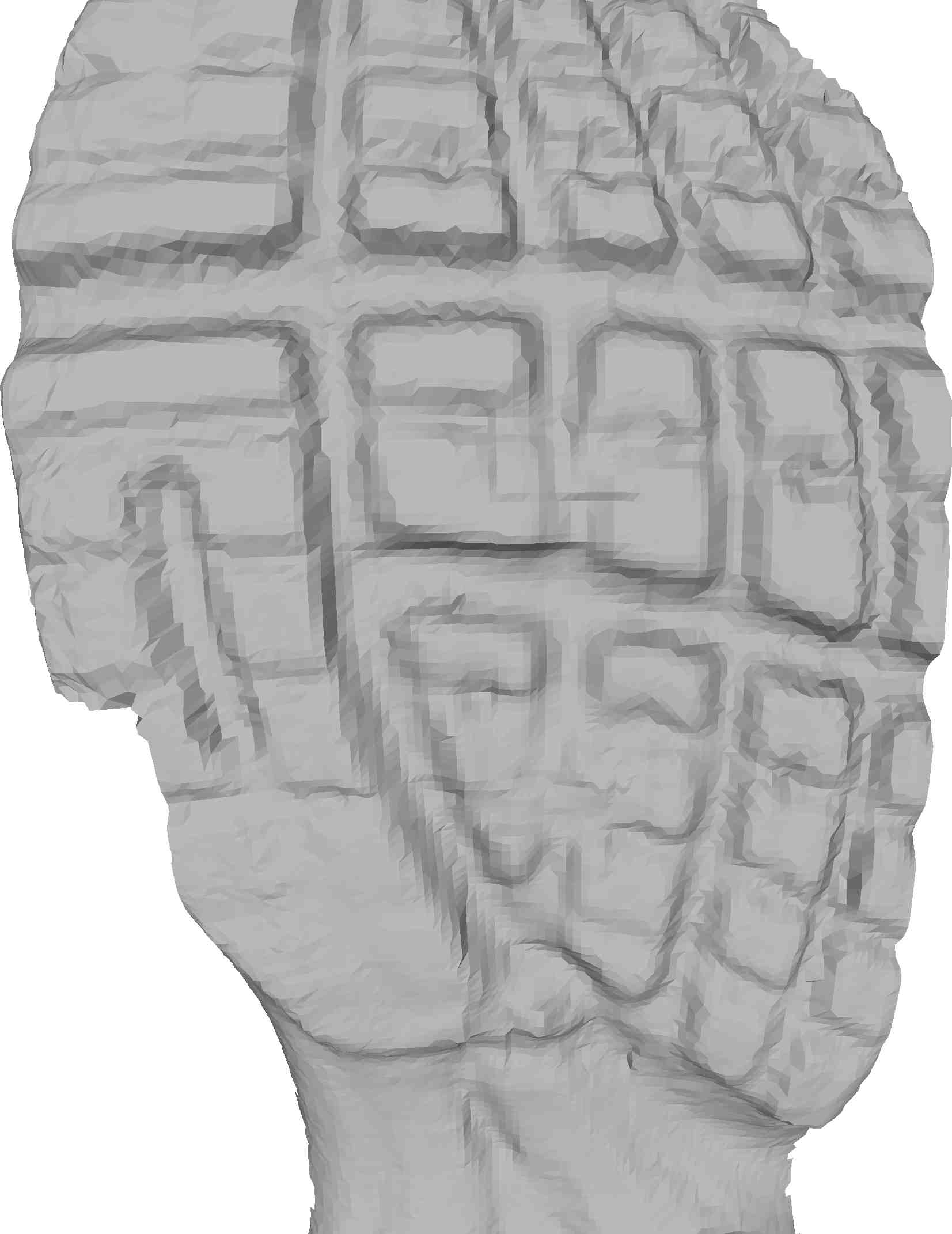}&
\includegraphics[width=\tablequalwidthWACV\linewidth]{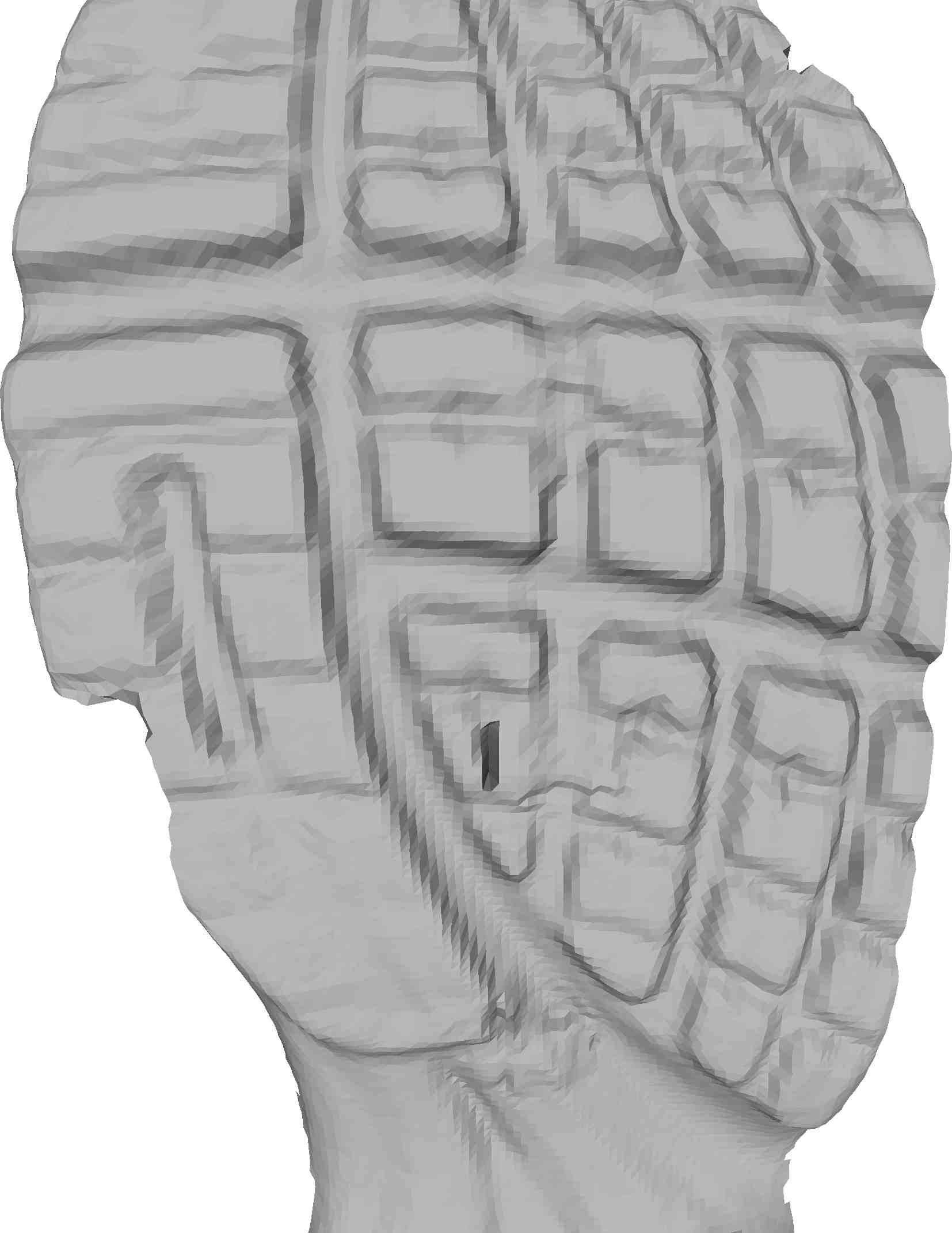}\\
\end{tabular}\begin{subfigure}{0.12\linewidth}
  \centering
  \includegraphics[width=1\linewidth]{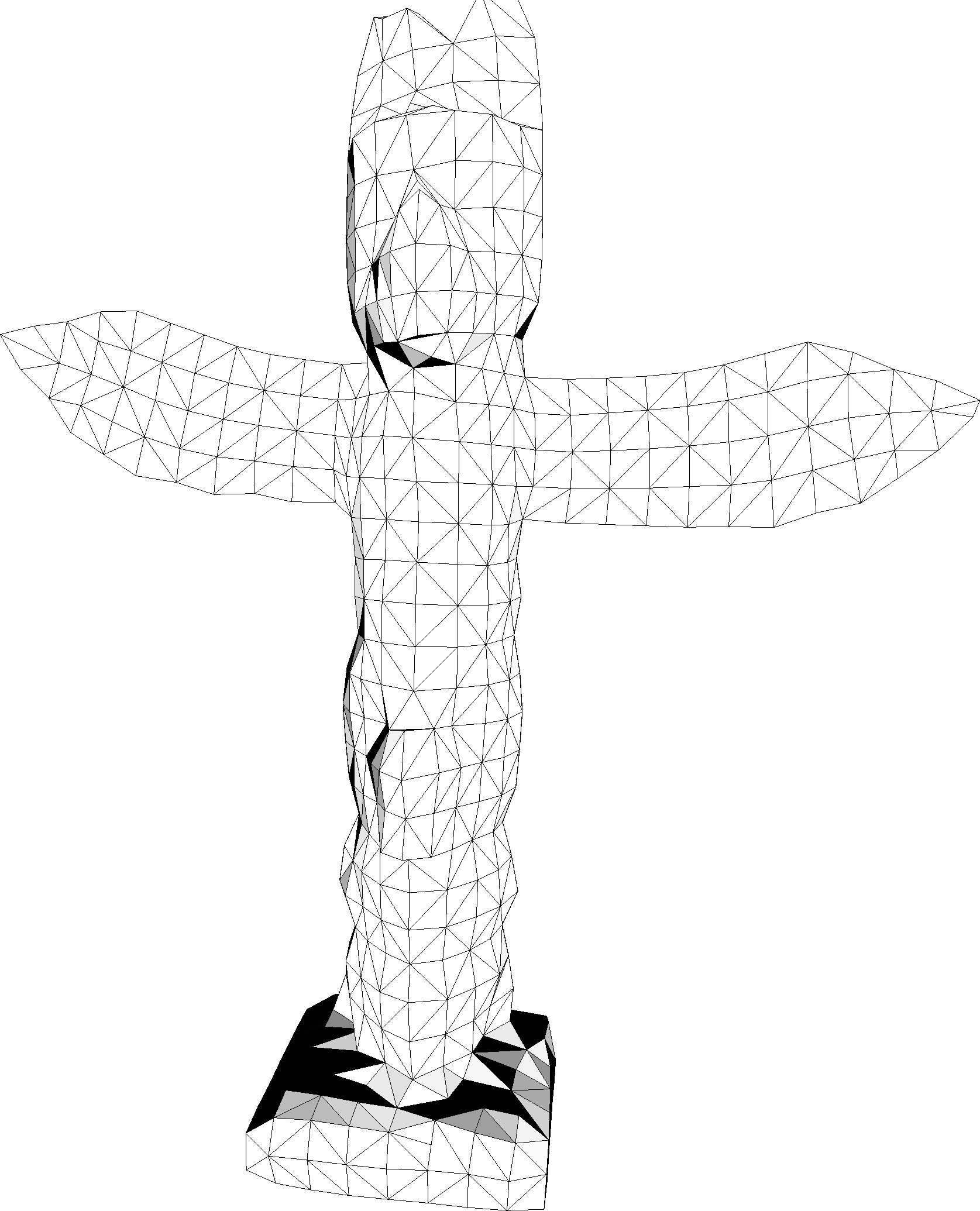}  
\end{subfigure}%
\begin{subfigure}{0.12\linewidth}
  \centering
  \includegraphics[width=1\linewidth]{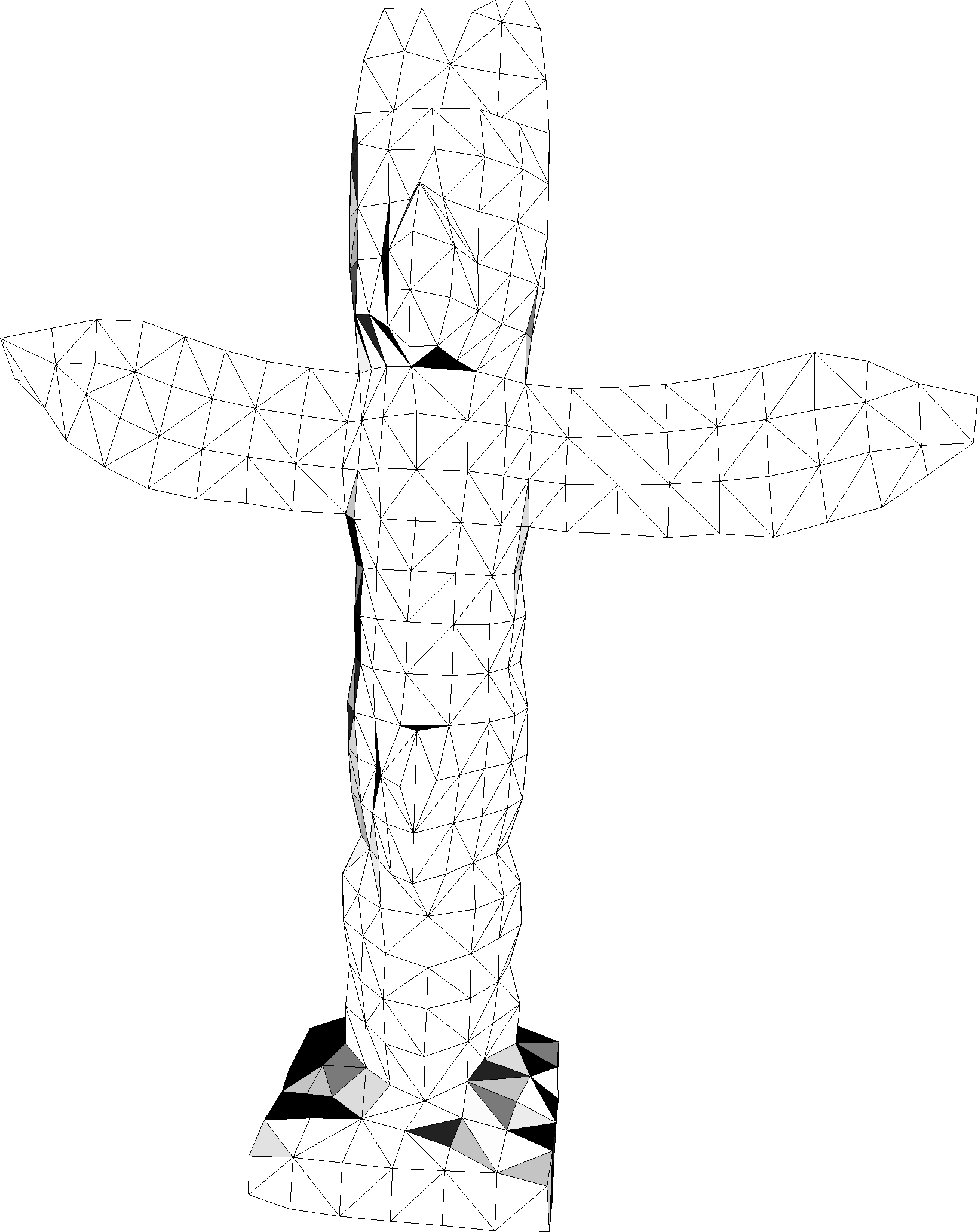}  
\end{subfigure}
\caption{(Left) Qualitative results of hole filling on the mesh Supernova with a hole radius of 0.025 with Global generative methods. (Right) Example of the quad mesh used in training (Left) and testing (Right) for the mesh Totem. Best viewed when zoomed digitally. Enlarged version and more results are provided in the supplementary material. }
\label{fig:inpaint_mesh_qual}
\end{figure*}

\begin{table}[t]
\centering
\begin{tabular}{lccc}
\toprule
Meshes & \cite{Liepa2003} & Local Dictionary&  Local CNN - small\_4x \\
\midrule
Supernova &  0.001646 &  0.000499 &  0.000415 \\
Terrex     &  0.001258 &  0.000595 &  0.000509 \\
Wander     &  0.002214 &  0.000948 &  0.000766 \\
LeatherShoe &  0.000854 &  0.000569 &  0.000512 \\
Brain      &  0.002273 &  0.000646 &  0.000457 \\

\bottomrule
\end{tabular}
\caption{Mean inpainting error of hole size 0.015, 0.025 and 0.035 for high texture dataset which uses Local patches generated on the same clean mesh of the corresponding shape.}
\label{table:inpaint_local}
\end{table}

\subsubsection{Hole filling on a single mesh}
\label{sec:singlemesh}
As explained before, our 3D patches from a single mesh are sufficient in amount to train a generative model for that mesh.
Note that we still need to provide an approximately correct scale for the quad mesh computation of the noisy mesh, so that the training and testing patches are not too different by size. 
Table \ref{table:inpaint_local} shows the result of hole filling using our smallest network - \textit{small\_4x} in terms of mean of the Cloud-to-Mesh error of the inpainted vertices and its comparison with our linear dictionary based inpainting results. We also provide the results from \cite{Liepa2003} in the table for better portray of the comparison. In this experiment, we learn one CNN per mesh on the patches in the clean input mesh (similar to local dictionary model), and tested in hole data as explained in the above section. As seen, our smallest network beats the result of linear approach of surface inpainting. 

We also train a long network \textit{long\_12x\_SC} (our best performing global network) with an offset factor of $k = 7$, giving us a total of 28 overlapping patches per quad location for the model \textit{Supernova} and we show the qualitative result in Figure \ref{fig:cnn_length} (Left). The figure verifies qualitatively, that with enough number of dense overlapping patches and a more complete CNN architecture, our method is able to inpaint surfaces with a very high accuracy.

\subsubsection{Global Denoising Autoencoder}
\label{sec:result_globalcnn}
Even though the input to the CNN are local patches, we can still create a single CNN designed for repairing a set of meshes, if the set of meshes are pooled from a similar domain. This is analogous to the \textit{global dictionary} where the dictionary was learnt from the patches of a pool of meshes. But to incorporate more variations in between the meshes in the set, the network needs to be well designed. 
This gets shown in the column \textit{global CNN} of Table \ref{table:inpaint_global} where our inpainting result with a single CNN (\textit{small\_4x}) for common meshes (Type 1 dataset) is comparable to our linear global-dictionary based method (column \textit{global dictionary}), but not better. With the premise that CNN is more powerful than the linear dictionary based methods, we perform additional experiments incorporating all the careful design choices discussed in the Section \ref{sec:networkdesign} for creating global CNNs for the purpose of inpainting different meshes in similar domain. The objective of these experiments are 1) to evaluate different denoising autoencoders ideas for inpainting in the context of height map based 3D patches 2) to verify that carefully designed generative CNNs performs better than the linear dictionary based methods, and 3) to show how to design a single denoising autoencoder for inpainting meshes from similar domain or inpainting meshes across a varied domain, when the number of meshes is not too high. We, however, do not claim that this procedure makes it possible to have a single model (be it global dictionary or global CNN) capable of learning and inpainting across a large number of meshes (say all meshes in ShapeNet); nor is this our intention. 

Figure \ref{fig:qualitative_test} provides the qualitative results for different networks showing the reconstructed patches from the masked incomplete patches. The results shows that the quality of the reconstruction increases with the increase in the network complexity. In terms of capturing overall details the network with FC layer seems to reconstruct the patches close to the original, but with the lack of contrast. This gets shown in the quantitative results where it is seen that the network with FC performs worse than most of networks. The quantitative results are shown in Table \ref{table:inpaint_local}. The best result qualitatively and quantitatively is shown by \textbf{long\_12x\_SC} - the longest network with symmetrical skip connections. Figure \ref{fig:cnn_length} (Right) provides more insights on the importance of the skip connections. Visualizations of the reconstructed hole filled mesh are provided in Figure \ref{fig:inpaint_mesh_qual} (Left).

\subsection{Generalisation capability}
\label{sec:generalization}
\textbf{Patches from common pool}
We perform reconstruction of \textit{Totem} using both the local dictionary and global dictionary having different number of atoms to know if the reconstruction error, or the shape information encoded by the dictionary, is dependent on where the patches come from at the time of training. We observed that when the number of dictionary atoms is sufficiently large (200 - 500), the global dictionary performs as good as the local dictionary (Figure \ref{fig:recerror_global_local} ). This is also supported by our superior performance of global dictionary in therms of hole filling. 

Keeping the number of atoms fixed at which the performances between Local and Global dictionary becomes indistinguishable (500 in our combined dataset), we learned global dictionary using the patches from different shapes, with one shape at a time. The reconstruction error of \textit{Totem} using these global dictionary varied very little. But we notice a steady increase in the reconstruction error with increase in the number of object used for learning; which becomes steady after a certain number of object. After that point (6 objects), adding more shapes for learning does not create any difference in the final reconstruction error (Figure \ref{fig:reconstructioncomplexity}). This verifies our hypothesis that the reconstruction quality does not deteriorate significantly with increase in the size of the dataset for common meshes for learning.

\textbf{Different test meshes}
We perform experiments to see how the inpainting method can be generalized among different shapes and use Type 1 dataset of \cite{Sarkar2017a} consisting of general shapes like Bunny, Fandisk, Totem, etc. These meshes do not have high amount of specific surface patterns. Column \textit{global CNN ex} of Table \ref{table:inpaint_global} shows the quantitative result for the network \textit{small\_4x} to inpaint the meshes trained on patches of other meshes. It is seen that if the shape being inpainted does not have too much characteristic surface texture, the inpainting method generalizes well. Note that this result is still better than the geometry based inpainting result  of \cite{Liepa2003}. Thus, it can be concluded that our system is a valid system for inpainting simple and standard surface meshes (Eg. \textit{Bunny}, \textit{Milk-bottle}, \textit{Fandisk} etc). 

However for complicated and characteristic surfaces (Eg. shoe dataset), we need to learn on the surface itself, because of the inherent nature of the input to our CNN - \textit{local patches} (instead of global features which takes an entire mesh as an input) that are supposed to capture surface details of its own mesh. Evaluating the generalizing capability of such a system requires patch computation on different locations between the training and testing set, instead of different mesh altogether. As explained before, in all our inpainting experiments, we explicitly made sure that the patches during the testing do not belong to training by manually computing a different set of quad mesh (Reference frames) for the hole triangulated mesh. To absolutely make sure the testing is done in a different set of patches, we manually tuned different parameters in \cite{Ebke2013} for quadriangulation. One example of such pair of quad meshes of the mesh Totem are shown in Figure \ref{fig:inpaint_mesh_qual} (Right).

The generalization capability can also be tested across the surfaces that are similar in nature, but from a different sample. 
The mesh Stone Wall from \cite{Zhou2013} provides a good example of such data, which has two different sides of the wall of similar nature. We fill holes on one side by training CNN on the other side and show the qualitative result in Figure \ref{fig:wall}. This verifies the fact that the CNN seems to generalize well for reconstructing unseen patches.

\textbf{Discussion on texture synthesis} We add a small discussion on the topic of texture synthesis as a good part of our evaluation is focused on a dataset of meshes high in textures.
As stated in the related work, both dictionary \cite{Aharon2006} based and BM3D \cite{Dabov2007} based algorithms are well known to work with textures in terms of denoising 2D images. Both approaches have been extended to work with denoising 3D surfaces. Because of the presence of patch matching step in BM3D (patches are matched and kept in a block if they are similar), it is not simple to extend it for the task of 3D inpainting with moderate sized holes, as a good matching technique has to be proposed for incomplete patches. Iterative Closest Point (ICP) is a promising means of such grouping as used by \cite{Rosman2013} for extending BM3D for 3D point cloud denoising. Since the contribution in \cite{Rosman2013} is limited for denoising surfaces, we could not compare our results with it - as further extending \cite{Rosman2013} for inpainting is not trivial and requires further investigation. Instead we compared our results with the dictionary based inpainting algorithm proposed in \cite{Sarkar2017a}.

\begin{figure}[t]
\centering
\begin{subfigure}{0.25\linewidth}
  \centering
  \includegraphics[width=\linewidth]{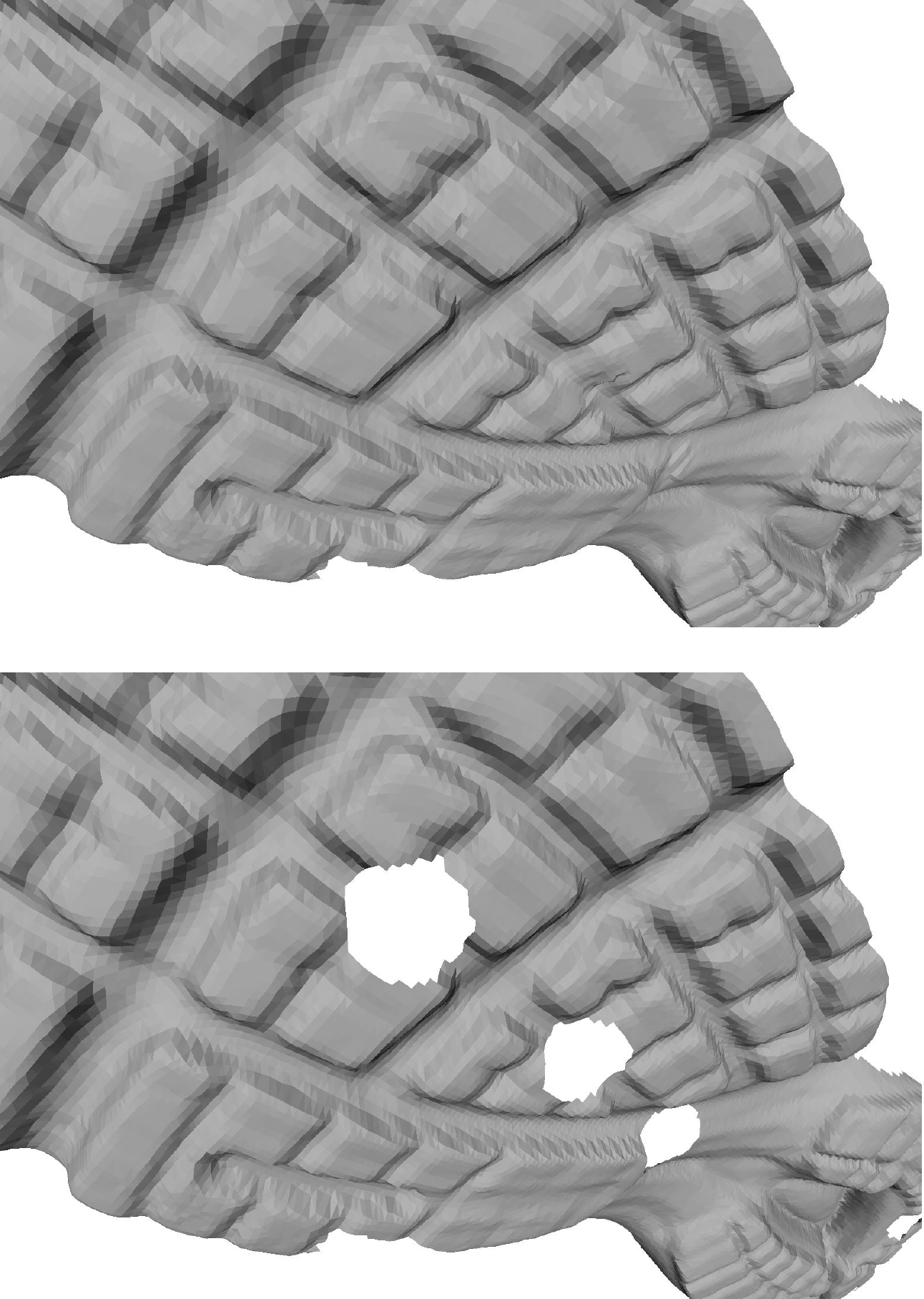}  
\end{subfigure} 
\begin{subfigure}{0.6\linewidth}
  \centering
  \includegraphics[width=\linewidth]{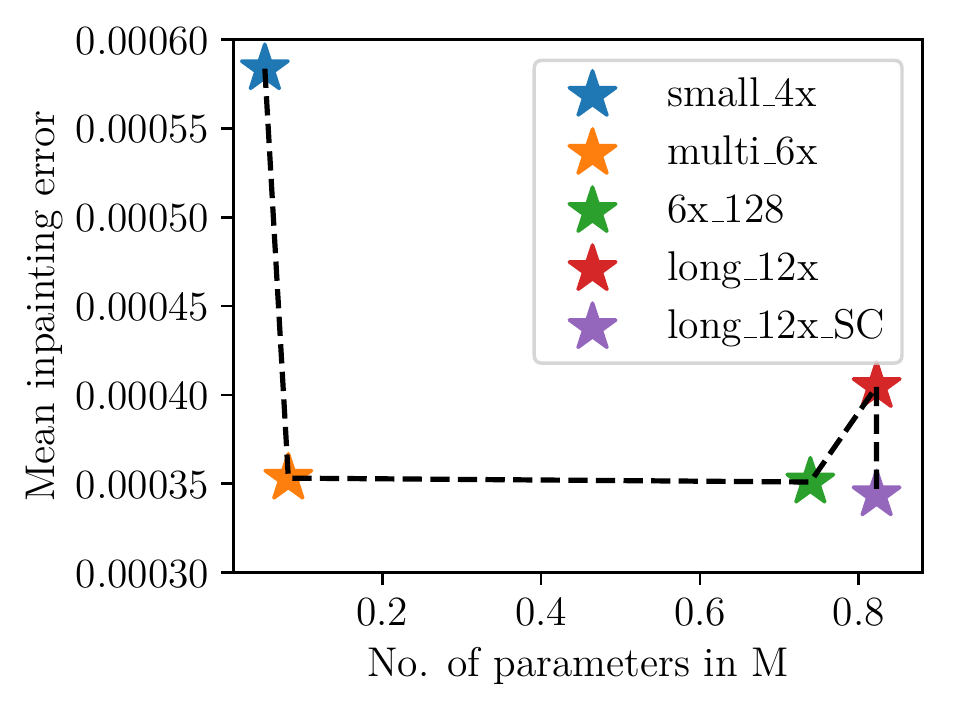}  
\end{subfigure}
\caption{(Left) Qualitative result of inpainting  on a single mesh with an overlap factor of $k = 7$. (Right) Mean inpainting error for high texture meshes wrt the number of parameters in the CNN. Inpainting error decreases with the increase in the network depth, saturates at one time, and performs worse if increased further. Presence of symmetrical skip connections decreases the error further providing its importance to train longer networks.}
\label{fig:cnn_length}
\end{figure}

Inpainting repeating structure is well studied in \cite{Pauly2008}. Because of the lack of their code and unavailability of results on a standard meshes, we could not compare our results to them. We also do not claim our method to be superior to them in high texture scenario, though we show high quality result with indistinguishable inpainted region for one of the meshes in Figure \ref{fig:cnn_length} (Left) using a deep network. However, we do claim our method to be more general, and to work in cases with shapes with no explicit repeating patterns (Eg. Type 1 dataset) which is not possible with \cite{Pauly2008}.

\begin{table}
\centering
\small
\begin{tabular}{lcc|c}
\toprule
{} &    global &    global CNN & global CNN ex\\
&   dictionary&  small\_4x & small\_4x \\
\midrule
Milk-bottle &     0.000123 &  0.000172 & 0.000187  \\
Baseball   &    0.000168 &  0.000113 & 0.000138\\
Totem      &    0.001052 &  0.001038 & 0.001406\\
Bunny      &   0.000569 &  0.000780 & 0.000644\\
Fandisk    &    0.000634 &  0.000916 & 0.000855\\
\bottomrule
\end{tabular}
\caption{(Left) Mean inpainting error of hole size 0.01, 0.02 and 0.03 for common mesh dataset using global models. For column \textit{global CNN} we use a single global CNN (small\_4x) trained on the local patches of all the meshes. The result of this small network is comparable to that of the linear global dictionary, but not better. This shows that we have more scope of improvement with a better network design for CNNs. 
(Right) in the column \textit{global CNN ex}, for each mesh, we use a global CNN (small\_4x) trained on the local patches of all the meshes except itself.  More discussion is in Section \ref{sec:generalization}
}
\label{table:inpaint_global}
\end{table}

\subsection{Limitation and failure cases}
\label{sec:failurecases}

\begin{figure}
\centering
\begin{subfigure}[b]{0.55\linewidth}
  \centering
  \includegraphics[width=1\linewidth]{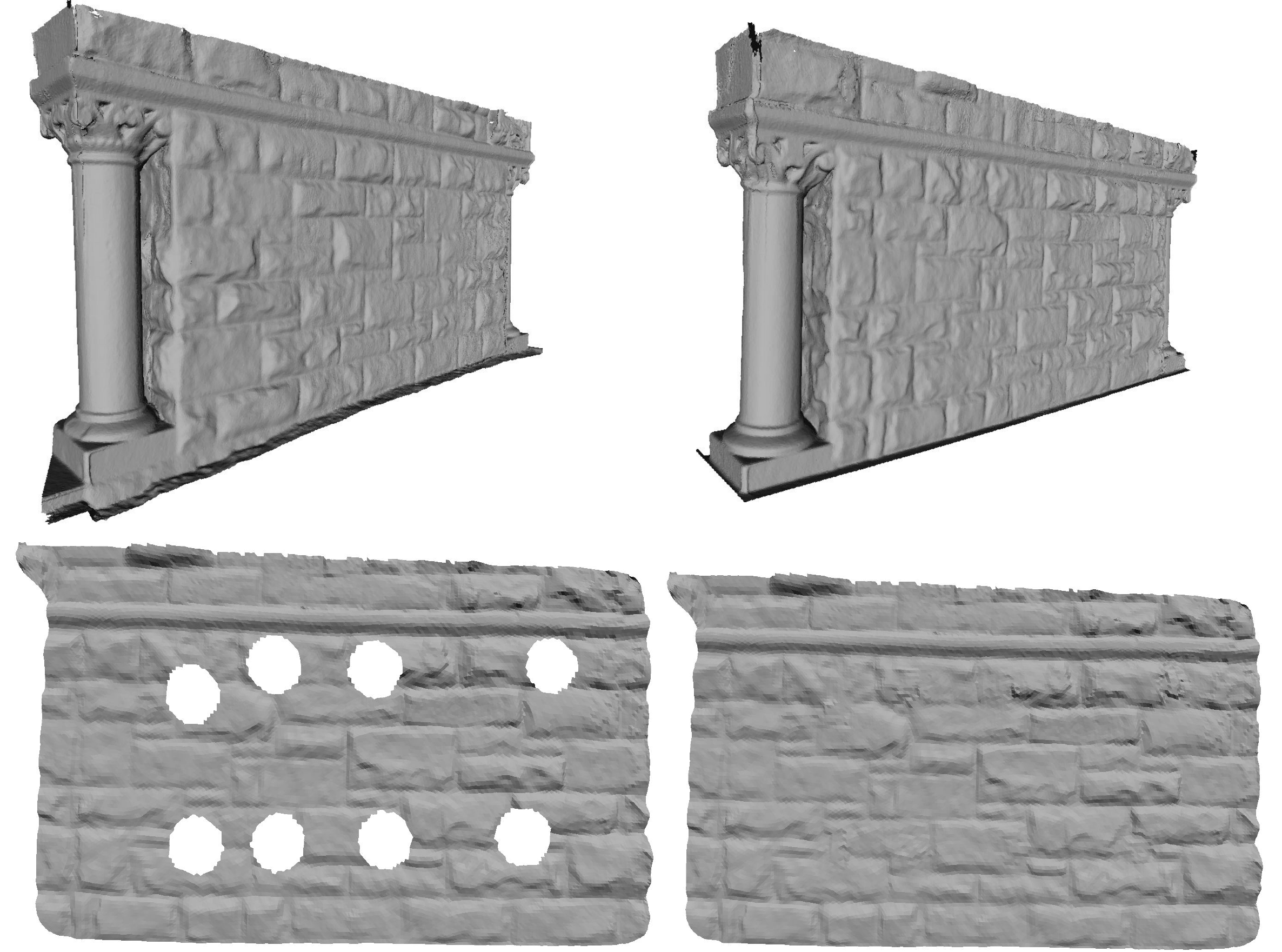} 
  \caption{Experiment on Stone Wall} 
  \label{fig:wall}
\end{subfigure}%
\begin{subfigure}[b]{0.45\linewidth}
  \centering
  \includegraphics[width=\linewidth]{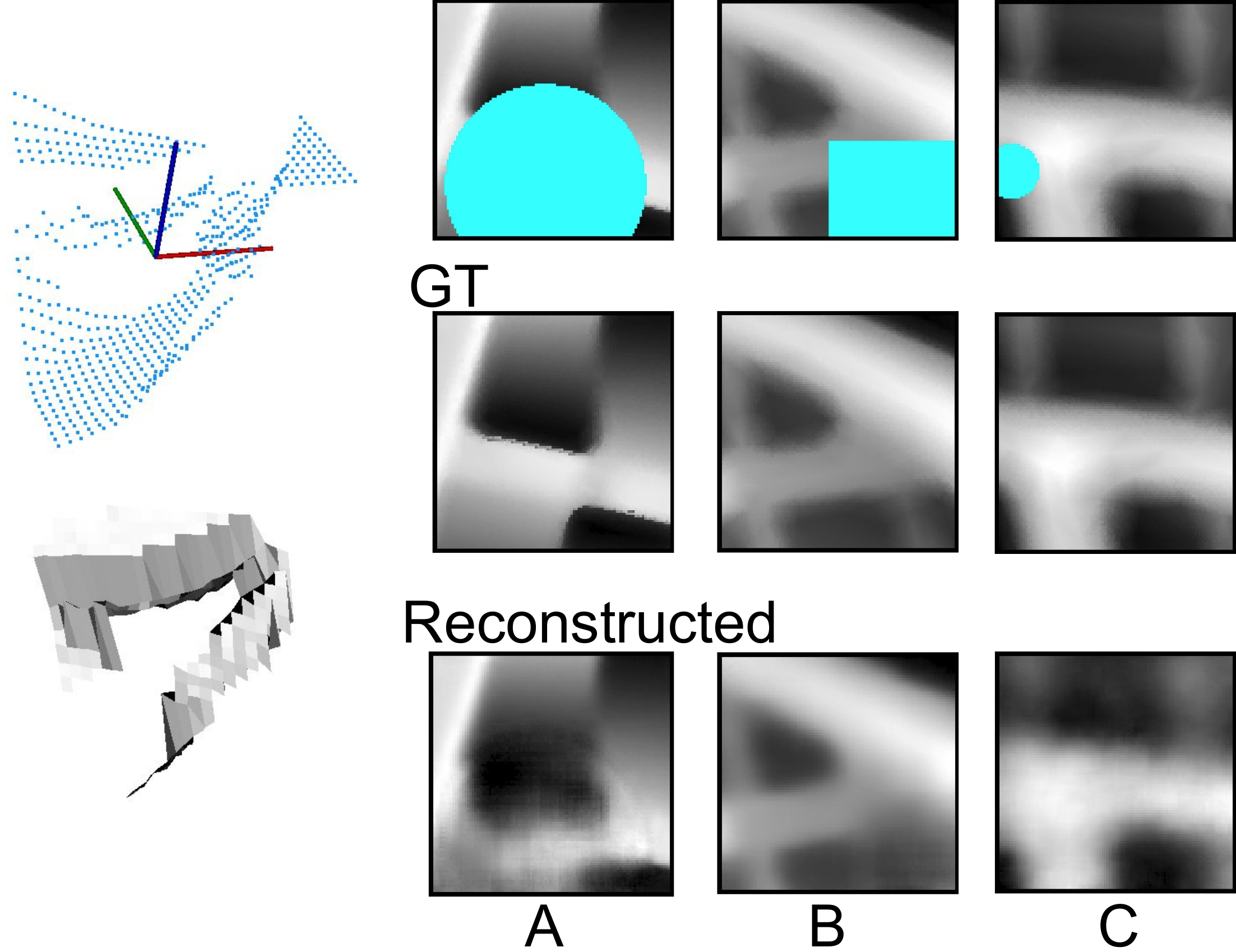}  
  \caption{Failure cases.}
  \label{fig:failurecase}
\end{subfigure}
\caption{(a) Scanned mesh of Stone Wall \cite{Zhou2013} which has two sides of similar nature shown in the top. The CNN \textbf{6x\_128} was trained on the patches generated on one side (Top Left) to recover the missing details on the other side (Top Right) whose result is shown in the bottom. 
(b) Failure cases -(Left) - bad or invalid patches (point cloud with RF at the top, and its corresponding broken and invalid surface representation at the bottom) at complicated areas of a mesh. (Right) Three failure case scenarios of the CNN.
}
\end{figure}

\noindent
\textbf{General limitations} - The quad mesh on the low resolution mesh provides a good way of achieving stable orientations for computing moderate length patch in 3D surfaces. However, on highly complicated areas such as joints, and a large patch length, the height map based patch description becomes invalid due to multiple overlapping surfaces on the reference quad as shown in Figure \ref{fig:failurecase} (left). Also, the method in general does not work with full shape completion where the entire global outline has to be predicted.

\noindent
\textbf{Generative network failure cases} - It is observed that small sized missing regions are reconstructed accurately by our long generative networks. Failure cases arise when the missing region is large. In the first case the network reconstructs the region according to the patch context slightly different than the ground truth (Figure \ref{fig:failurecase}-A). The second case is similar to the first case where the network misses fine details in the missing region, but still reconstructs well according to the other dominant features. The third case, which is often seen in the network with FC, is the lack of contrast in the final reconstruction (Figure \ref{fig:failurecase}-C). Failure cases for smaller networks can be seen in Figure \ref{fig:qualitative_test}.

\section{Conclusion}
\label{sec:conclusion}

We proposed in this paper our a first attempt at using generative models on 3D shapes with a representation and parameterization other than voxel grid or 2D projections. For that, we proposed a new method for shape encoding 3D surface of arbitrary shapes using rectangular local patches.
With these local patches we designed generative models, inspired that from 2D images, for inpainting moderate sized holes and showed our results to be better than the geometry based methods. 
With this, we identified an important direction of future work - exploration of the application of CNNs in 3D shapes in a parameterization different from the generic voxel representation. 
In continuation of this particular work, we would like to extend the local quad based representation to global shape representation which uses mesh quadriangulation, as it inherently provides a grid like structure required for the application of convolutional layers. This, we hope, will provide an alternative way of 3D shape processing in the future.

\ifCLASSOPTIONcaptionsoff
  \newpage
\fi



%

%

\bibliographystyle{IEEEtran}
\bibliography{vision_general}

%








\end{document}